\documentclass[10pt,journal,compsoc]{IEEEtran}
\usepackage{graphicx}

\ifCLASSOPTIONcompsoc
  \usepackage[nocompress]{cite}
\else
  \usepackage{cite}
\fi
\usepackage{hyperref}       
\usepackage{url}   

\usepackage{amsmath,amsfonts,bm}

\def\eqref#1{equation~\ref{#1}}

\def\1{\bm{1}}

\def\vg{{\bm{g}}}

\def\vr{{\bm{r}}}

\def\vu{{\bm{u}}}
\def\vv{{\bm{v}}}

\def\vx{{\bm{x}}}
\def\vy{{\bm{y}}}
\def\vz{{\bm{z}}}

\def\mS{{\bm{S}}}

\def\mV{{\bm{V}}}

\DeclareMathAlphabet{\mathsfit}{\encodingdefault}{\sfdefault}{m}{sl}
\SetMathAlphabet{\mathsfit}{bold}{\encodingdefault}{\sfdefault}{bx}{n}

\def\gL{{\mathcal{L}}}

\newcommand{\E}{\mathbb{E}}

\usepackage{amsthm}
\theoremstyle{plain}
\newtheorem{theorem}{Theorem}[section]

\theoremstyle{definition}

\theoremstyle{remark}

\usepackage{epsfig}
\usepackage{graphicx}
\usepackage{booktabs}       
\usepackage{amsfonts}       
\usepackage{nicefrac}       
\usepackage{microtype}      
\usepackage{caption}
\usepackage[table]{xcolor}         
\usepackage{amsmath}
\usepackage{amssymb}
\usepackage{cleveref}
\usepackage{multirow}
\usepackage{enumerate}
\usepackage{cancel}
\usepackage{makecell}
\usepackage{algorithm}
\usepackage{algorithmic}
\usepackage{enumitem}
\usepackage{diagbox}
\usepackage{listings}
\usepackage{soul}
\usepackage{tabularx}
\definecolor{codegreen}{rgb}{0,0.6,0}
\definecolor{codegray}{rgb}{0.5,0.5,0.5}
\definecolor{codepurple}{rgb}{0.58,0,0.82}
\definecolor{backcolour}{rgb}{0.95,0.95,0.92}

\lstdefinestyle{mystyle}{
    backgroundcolor=\color{backcolour},   
    commentstyle=\color{codegreen},
    keywordstyle=\color{magenta},
    numberstyle=\tiny\color{codegray},
    stringstyle=\color{codepurple},
    basicstyle=\ttfamily\footnotesize,
    breakatwhitespace=false,         
    breaklines=true,                 
    captionpos=b,                    
    keepspaces=true,                 
    numbers=left,                    
    numbersep=5pt,                  
    showspaces=false,                
    showstringspaces=false,
    showtabs=false,                  
    tabsize=2
}
\setlist[itemize]{leftmargin=*}
\ifCLASSINFOpdf
\else
\fi

\usepackage{pifont}
\newcommand{\cmark}{\text{\ding{51}}}%
\newcommand{\xmark}{\text{\ding{55}}}%

\def\mS{{\mathbf{S}}}

\def\mV{{\mathbf{V}}}

\begin{document}
%
\title{Unsupervised Representation Learning from Sparse Transformation Analysis}
%
%
%
\author{Yue~Song,
        T.~Anderson~Keller,
        Yisong~Yue,
        Pietro~Perona,
        Max~Welling
\IEEEcompsocitemizethanks{\IEEEcompsocthanksitem Yue Song, Yisong Yue, Pietro Perona are with Computing and Mathematical Sciences, Caltech, CA. T. Anderson Keller is with the Kempner Institute for the Study of Natural and Artificial Intelligence, Harvard University, MA. Max Welling is with Amsterdam Machine Learning Lab, University of Amsterdam, the Netherlands. \\
E-mail: \{yuesong, yyue, perona\}@caltech.edu, takeller@fas.harvard.edu, m.welling@uva.nl}
\thanks{Manuscript received April 19, 2005; revised August 26, 2015.}}

%
%

\markboth{IEEE TRANSACTIONS ON PATTERN ANALYSIS AND MACHINE INTELLIGENCE}%
{Shell \MakeLowercase{\textit{et al.}}: Bare Demo of IEEEtran.cls for Computer Society Journals}
%



\IEEEtitleabstractindextext{%
\begin{abstract}
  There is a vast literature on representation learning based on principles such as coding efficiency, statistical independence, causality, controllability, or symmetry. In this paper we propose to learn representations from sequence data by factorizing the transformations of the latent variables into sparse components. Input data are first encoded as distributions of latent activations and subsequently transformed using a probability flow model, before being decoded to predict a future input state. The flow model is decomposed into a number of rotational (divergence-free) vector fields and a number of potential flow (curl-free) fields. Our sparsity prior encourages only a small number of these fields to be active at any instant and infers the speed with which the probability flows along these fields. Training this model is completely unsupervised using a standard variational objective and results in a new form of disentangled representations where the input is not only represented by a combination of independent factors, but also by a combination of independent transformation primitives given by the learned flow fields. When viewing the transformations as symmetries one may interpret this as learning approximately equivariant representations. Empirically we demonstrate that this model achieves state of the art in terms of both data likelihood and unsupervised approximate equivariance errors on datasets composed of sequence transformations.
\end{abstract}

}

\maketitle

\IEEEdisplaynontitleabstractindextext

\section{Introduction}

The resounding success of deep learning in the last decade has largely been attributed to the ability of deep neural networks to learn valuable internal representations directly from data. Such representations are now at the forefront of many of today's most advanced technologies, allowing for the extraction of abstract semantics from high dimensional data, and enabling previously unimaginable technologies such as automatic image inpainting and apparent natural language understanding. Although these impressive affordances of representation learning are only very recently showing their true potential, there is an extensive history of work searching for a consensus on what are the ultimate principles which define a `good' representation.

\begin{figure}[tbp]
    \centering
    \includegraphics[width=0.99\linewidth]{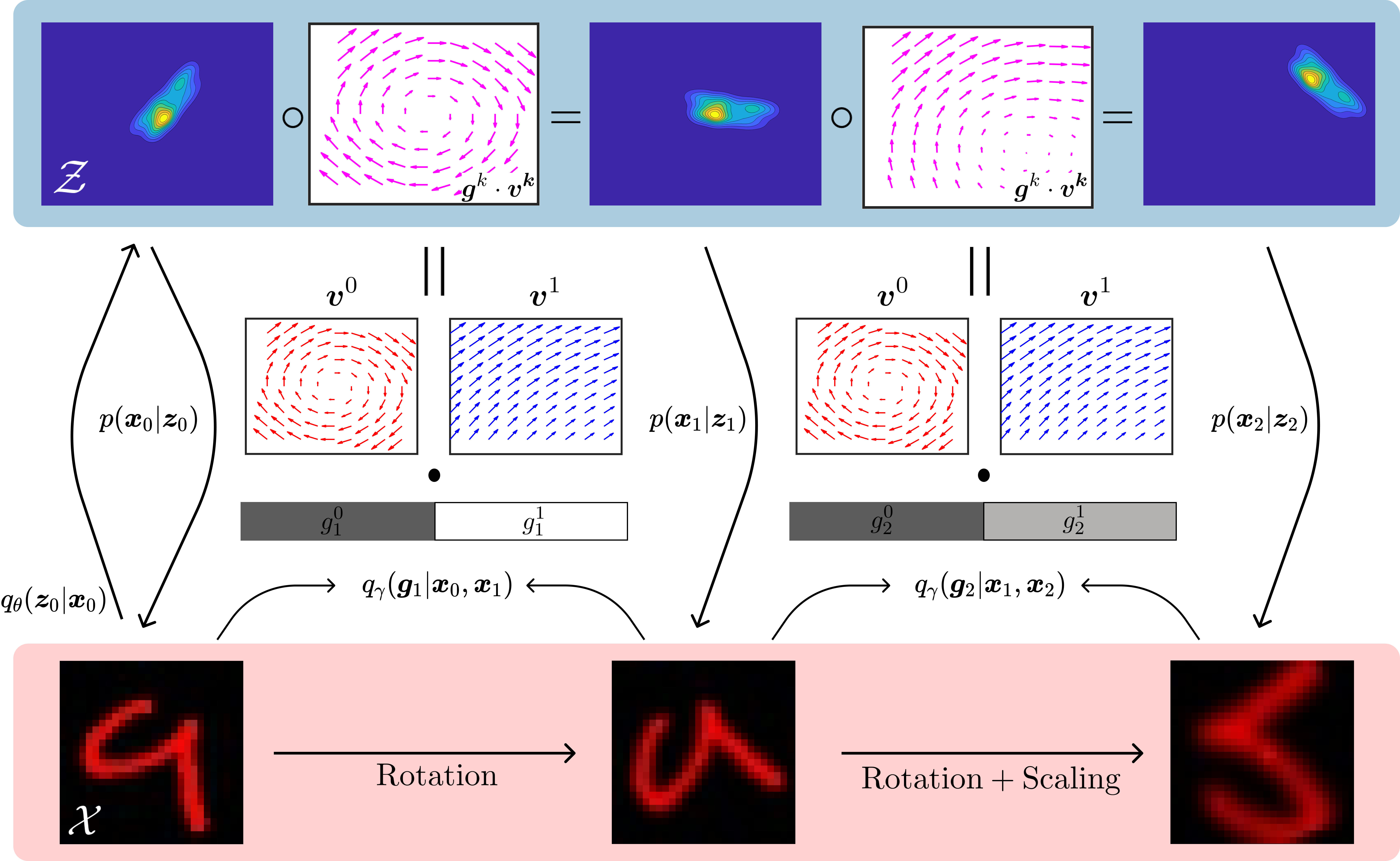}
    \caption{Overview of Sparse Transformation Analysis (STA): given an input sequence $\{\mathbf{x}_t\}_t^T$ containing some unknown combination of transformations, the model infers both an initial distribution over latent variables $q_{\theta}(\mathbf{z}_0|\mathbf{x}_0)$, and a sparse coefficient vector for each time-step $\vg_t  = \{g_t^0, g_t^1, \ldots, g_t^K\}$ which is used to linearly combine the K distinct learned vector fields $\mathbf{v}^k = (\nabla u^k + \vr^k)$ in order to compute the latent transformation from $\vz_t$ to $\vz_{t+1}$. Due to the sparsity of $\vg$, the model learns to disentangle the observed transformations into separate flow fields $\vv^k$ entirely unsupervised.
    }
    \label{fig:teaser}
\end{figure}

One early line of work in representation learning focused on ideas of redundancy reduction, believing that biological neural systems would naturally strive for an efficient code due to competitive pressures \cite{barlow1961possible}. Building on this idea, the principles of sparsity and statistical independence of coding dimensions emerged as guidelines for learning such maximally efficient codes, eventually resulting in the frameworks of sparse coding \cite{olshausen1997sparse} and independent component analysis \cite{comon1994independent}. 
Inspired by the fact that natural intelligence is embedded in a world where physical laws restrict observations to sequences of smooth transformations, these ideas of efficiency and sparsity were extended to include temporal dimensions. A seminal example is Slow Feature Analysis \cite{wiskott2002slow}, a learning framework which assumes that individual latent variables are likely to change slowly over time. Models adhering to these principles were shown to learn invariances directly from data and uncover underlying generative factors if those factors had similar slow dynamics. Recent work has further shown that natural videos follow a specific sparse transition structure, meaning that the set of generative factors which describe a given input sequence is mostly constant over time with sparse transitions between which factors are active. Klindt \emph{et al.}~\cite{klindt2021towards} then demonstrated that by building a model which incorporates this structure into its prior, it is possible to provably learn the true generative factors of video data in an unsupervised manner. 
While differing in implementation and methodology, all these frameworks appear in some sense to share the goals of learning meaningful, interpretable, `disentangled', and controllable latent codes such that specific directions in the latent space corresponded to the independent factors which were responsible for generating the input data distribution. 

More recently, the concept of equivariance has emerged as a mathematical framework for learning highly structured and thereby `controllable' latent representations \cite{cohen2016group}. Specifically, equivariant neural networks are built to explicitly respect the symmetries of the input domain in their output space. In such models, there are known predictable output transformations for given input transformations of interest. To date, these concepts form the foundation of some of the most precisely formalized definitions of `disentanglement' in the literature \cite{higgins2018towards, cohen2015transformation}. In prior work however, there is a relatively sharp divide between equivariant neural networks and models which are focused on disentanglement. Specifically, it is currently only known how to build networks which are equivariant with respect to known transformations which have a mathematical group structure. This includes traditional coordinate symmetries \cite{cohen2017steerable, hoogeboom2022equivariant}, but is severely limiting when considering the types of natural image transformations that are typically explored in disentangled representation learning. One line of research has aimed to bridge this divide by learning `approximately equivariant' models which are intended to learn these types of structured representations directly from data itself \cite{keller2021topographic, song2023flow}. However the vast majority of the models in this domain require at least some form of weak supervision of segmented sequences with only single transformations being observed. 

In this paper, we introduce a new modeling framework, denoted Sparse Transformation Analysis (STA), which takes inspiration from these foundational representation learning approaches, thereby yielding what we argue to be a uniquely structured yet flexible latent space which aligns with natural data statistics. Fig.~\ref{fig:teaser} depicts the overview of our STA. The framework requires no supervision of input sequences, assuming only that the observed transformations from one timestep to the next match a sparse transition structure similar to that observed by \cite{klindt2021towards}. Specifically, STA takes a generative modeling approach, asserting that generative factors should be represented by distributions over latent variables, and that these distributions should flow smoothly in the latent space in concert with the smooth flow of observations in the world. Furthermore, the framework posits that this flow should not be arbitrary, but can be represented as a sparse combination of learned flow field primitives. In alignment with notions of disentanglement and approximate equivariance, these flow field primitives can be seen as directions in latent space which correspond to observed input transformations. Unlike \cite{klindt2021towards}, STA allows for highly flexible latent dynamics for each transformation `direction' by parameterizing each transformation's flow field through the Helmholtz decomposition as a combination of curl-free and divergence-free components. The specific sparse combination of flow fields which are used to transform the latent distribution from one time-step to the next is treated as an unobserved latent variable with a multivariate history-dependent spike and slab prior~\cite{mitchell1988bayesian}, and inferred simultaneously with the other latent variables through amortized variational inference. In this way, this approach can be seen to combine ideas of sparse coding, slow feature analysis, and approximate equivariance, while still allowing deep neural network feature extractors to be leveraged in a relatively unconstrained manner.

In the following, we will demonstrate that this framework yields the state of the art in unsupervised approximate equivariance, as quantified through a measured equivariance error, and further that our method yields the highest likelihood on the test set in the unsupervised setting. As desired, we additionally observe that the model automatically learns to separate observed transformations into independent flow fields, and that these latent flows can be flexibly combined or switched during traversal. By tuning the magnitude of these flow fields, our model also has precise control of the transformation speed. Further, when slightly modifying the spike component to two separate controls, our method can learn to segregate latent symmetries and invariances into the two distinct components of the vector fields. The decomposition of latent flow fields is coherent with the categories of input transformations. Besides simple experiments on toy datasets, we further validate our STA on use cases of real-world video analysis, including movements of robot arms~\cite{nie2020semi}, lighting changes of indoor scenes~\cite{nie2020semi}, behavior videos of social agents~\cite{sun2021multi}, and ego-centric autonomous driving videos~\cite{cordts2016cityscapes}. Our method can identify a wide range of independent motions in the video sequences. Ultimately, we present this model as a natural next step in the development of unsupervised approximately equivariant representation learning algorithms. Code is publicly available at \href{https://github.com/KingJamesSong/latent-flow}{https://github.com/KingJamesSong/latent-flow}.

This paper is an extension of \cite{song2023flow}. We build our model upon \cite{song2023flow} in the high-level concept of using latent flow fields for modeling transformations, but we do have two substantial modifications: (1) our method uses spike-and-slab prior to avoid any sort of supervision; (2) we leverage Helmholtz decomposition for more expressive/flexible latent flows. These improvements could bring many concrete benefits. For example, the elimination of supervision greatly broadens the applicability of the method, thus making it applicable to real-world video understanding (\emph{e.g.,} CalMS~\cite{sun2021multi} and Cityscape~\cite{cordts2016cityscapes} in Sec.~\ref{sec:real_video}). Further, the slab component mimics the motion speed in natural videos and allows for the explicit control of transformation speeds, which is seldom studied in the literature of disentangled representation learning. Moreover, as discussed in Sec.\ref{sec:sep_control}, Helmholtz decomposition allows each transformation to be associated with either a curl-free or divergence-free component, thereby offering improved interpretability of the learned transformation structure. Finally, to complement the empirical improvements, we also provide a formal identifiability argument grounded in sparse dictionary learning, as detailed in Sec. B of the supplementary material.
\begin{figure*}[thbp]
    \centering
    \includegraphics[width=0.9\linewidth]{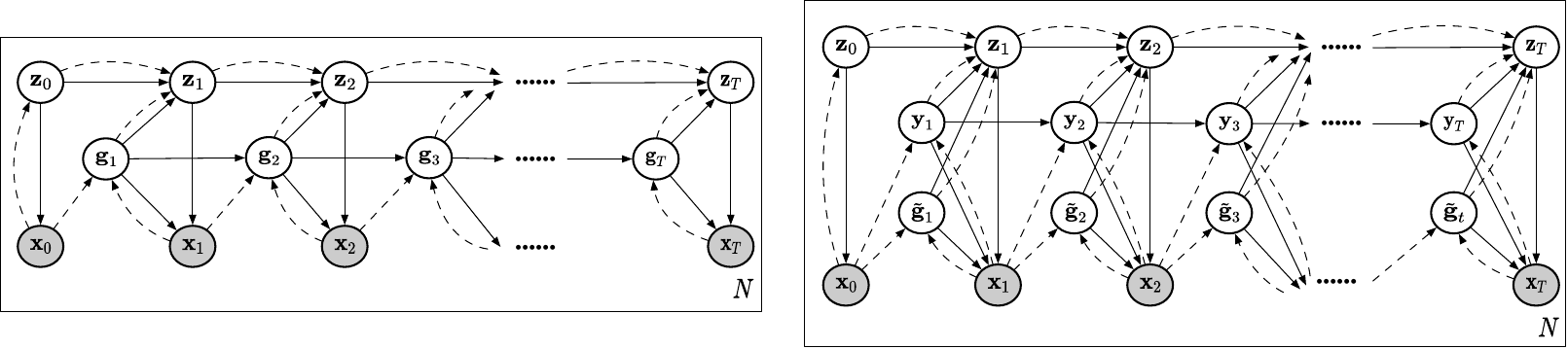}
    \caption{Our model across $N$ sequences in plate notation (Left) and a detailed version with decomposed spike and slab components (Right). White nodes denote latent variables, shaded nodes denote observed variables, solid lines denote the generative model, and dashed lines denote the approximate posterior. Different from the spike component $\vy_t$, the slab variable $\Tilde{\vg}_t$ is independent across timesteps.}
    \label{fig:graph_model}
\end{figure*}

\section{Related Work}

\subsection{Disentangled/Equivariant Representation Learning} 

Disentanglement and equivariance are considered two important desiderata of representation learning~\cite{higgins2018towards}. The idea of learning disentangled representation was first studied by InfoGAN~\cite{chen2016infogan} and $\beta$-VAE~\cite{higgins2016beta}. InfoGAN~\cite{chen2016infogan} disentangles the latent space by maximizing the mutual information between a subset of latent dimensions and observations, while $\beta$-VAE~\cite{higgins2016beta} factorizes the posterior $q(\mathbf{z}|\mathbf{x})$ by penalizing the total correlation between the prior and variational posterior. Subsequent work following InfoGAN mainly focused on discovering different semantically interpretable directions in the latent space~\cite{goetschalckx2019ganalyze,jahanian2020steerability,voynov2020unsupervised,harkonen2020ganspace,zhu2020learning,peebles2020hessian,shen2021closed,wei2021orthogonal,zhu2021low,Tzelepis_2021_ICCV,zhu2022region,song2022orthogonal,oldfield2022panda,song2023latent,song2023householder}. Following $\beta$-VAE, many attempts have been made to encourage independence of the aggregated posterior through additional guidance~\cite{dilokthanakul2017deep,dupont2018learning,kumar2018variational,kim2018disentangling,chen2018isolating,jeong2019learning,yildiz2019ode2vae,ding2020guided,shao2020controlvae,locatello2020weakly,tai2022hyperbolic,estermann2023dava}. More recently, one line of research proposed to disentangle diffusion models by crafting compact low-dimensional latent spaces~\cite{kwon2022diffusion,park2023understanding,yang2023disdiff,wang2023infodiffusion}. Parallel developments in causal inference literature emphasize that disentanglement can be viewed as identifying independent causal mechanisms~\cite{spirtes2000causation,pearl2009causality,locatello2019challenging,yang2021causalvae}. Ke~\emph{et al.}~\cite{ke2019learning} demonstrated the potential for neural networks to learn causal models by interpreting unknown interventions. Kumar and Sinha~\cite{kumar2021disentangling} extended this idea by proposing methods for disentangling mixtures of unknown causal interventions, further bridging representation learning and causal inference. More recently, Song~\emph{et al.}~\cite{song2024causal} explored causal temporal representation learning, explicitly modeling nonstationary and sparse causal transitions in temporal data, thus advancing understanding of disentangled representations in dynamic scenarios.

Equivariant networks, on the other hand, are usually more strictly defined than disentanglement methods. Analytical approaches typically enforce neural network weights to explicitly respect the symmetry of group transformations~\cite{cohen2016group,cohen2017steerable,ravanbakhsh2017equivariance,worrall2017harmonic,worrall2019deep,finzi2020generalizing,hoogeboom2022equivariant}. However, as noted in the introduction, the transformation for which such analytic equivariance is possible is limited to certain groups like special orthogonal groups, and may not apply to real-world scenarios. To avoid this issue, a number of recent models have aimed to relax this constraint and instead learn approximately equivariant representations directly from data~\cite{diaconu2019learning,connor2021variational,klindt2021towards,keller2021topographic,song2023latent,song2023flow}. 

\subsection{Sequential Disentanglement}

Another closely related research branch of disentangled representations is sequential disentanglement~\cite{hsu_unsupervised2017,denton2017unsupervised,villegas2017decomposing,lisevae2018,tulyakov2018mocogan,zhu2020s3vae,bhagat2020disentangling,yamada2020disentangled,bai2021contrastively,han2021disentangled,tonekaboni2022decoupling,naiman2023operator,berman2023multi,nimrod2024sequential} where the disentangled representation learning techniques are applied to sequence data like video and audio. In the sequential case, latent variables are typically split into single static time-invariant codes that do not change over time and multiple dynamic time-varying components that describe the distinct motions in the sequence. Due to the static and dynamic assumptions, these methods have to use two sets of latent variables for modeling different components. Differently, we assume that the static identity information is given in the latent variable, and the dynamic sequential transformations are encoded in the latent flow fields. Further, these approaches achieve disentanglement of single latent dimensions through implicit KL regularization. By contrast, we leverage the sophisticated sparsity constraints to explicitly classify the transformations and factorize them into different latent flows.

\subsection{Physical Inductive Biases in Deep Learning} 

The performance of deep learning models is heavily based on inductive biases. In recent years, an increasing amount of effort has developed to endow deep neural networks with physical priors and inductive biases (\emph{e.g.,} symmetries or conservation laws). Much attention has been focused on using neural networks to solve Partial Differentiable Equations (PDEs), such as Physics Informed Neural Networks (PINNs)~\cite{pinns} and other improved variants~\cite{hsieh2019learning,brandstetter2022message,richter2022neural,zeng2023competitive,bajaj2023recipes,akhound2023lie}. Other active research directions include handling input symmetries with aforementioned equivariant networks, building generative score-based denoising diffusion models using Fokker-Planck equations~\cite{ho2020denoising,song2020denoising,song2020score}, and designing neural networks with Hamiltonian dynamics for improved generalization~\cite{greydanus2019hamiltonian,toth2019hamiltonian}. In this work, we leverage PINNs to place constraints on our latent flow fields such that they obey the assumptions of fluid-dynamic optimal transport and the Helmholtz decomposition, thereby increasing the expressivity while including valuable inductive biases.
\section{The Generative Model}

This section introduces the probabilistic framework of our generative model. We start with the factorization of sequence distributions, followed by the spike and slab priors, and end with the time evolution of the latent priors.

\begin{figure}[htbp]
    \centering
    \includegraphics[width=0.99\linewidth]{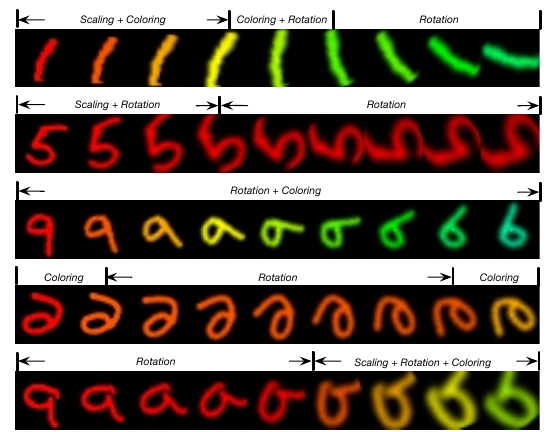}
    \caption{Exemplary sequences generated by our spike prior.}
    \label{fig:sequence}
\end{figure}

\subsection{Factorized Sequence Distributions}

Fig.~\ref{fig:graph_model} depicts the plate diagram of our model through solid lines. As can be seen, our model defines a distribution over $N$ sequences of observed variables $\bar{\vx} = \{\vx_0, \vx_1 \ldots, \vx_T\}$. The sequence distribution is factorized into $K$ distinct basic components as we assume each observed sequence is generated by the linear combination of $K$ separate basis flows in latent space. To model the discrete sequences of observations, we aim to define a joint distribution with a similarly discrete sequence of latent variables $\bar{\vz} = \{\vz_0, \vz_1 \ldots, \vz_T\}$ describing the observations, and $\bar{\vg} = \{\vy_1\cdot\Tilde{\vg}_1, \vy_2\cdot\Tilde{\vg}_2 \ldots, \vy_T\cdot\Tilde{\vg}_T\}$ describing the transformation \textit{type} ($\vy_t$) and \textit{speed} ($\Tilde{\vg}_t$) happening between neighboring observations. Specifically, we assert the following factorization of the joint distribution over $T$ timesteps:
\begin{equation}
\label{eqn:joint_p}
\begin{gathered}
p(\bar{\vx}, \bar{\vz}, \bar{\vg}) = p(\vg_1)p(\vz_0)p(\vx_0|\vz_0)\\ \prod_{t=1}^T p(\vz_t | \vz_{t-1}, \vg_t) p(\vx_t | \vz_t, \vg_t)p(\vg_{t+1}|\vg_t).    
\end{gathered}
\end{equation}
Here $p(\vz_0)$ is a standard Normal distribution, $p(\vx_t|\vz_t, \vg_t)$ asserts a mapping from latents to observations, and $p(\bar{\vg})$ is the sequence of the random variables that controls the temporal variations of the transformation type and speed.

\subsection{Spike and Slab Priors} 
\label{sec:sp_sl}

We model real-world video as a sparse combination of transformation primitives.  To model this transition sparsity, we impose a spike and slab prior~\cite{mitchell1988bayesian} on the transformation variable $\vg_t$ for generating the sequences. The distribution is factorized as follows:
\begin{equation}
    p(\vg_t) = p(\vy_t) p(\Tilde{\vg}_t)
\end{equation}
where the `spike' variable $\vy_t$ is a multi-hot vector that selects the specific transformation primitives to combine, and the `slab' variable $\Tilde{\vg}_t$ controls the transformation speed. The spike component usually concentrates its mass around zero, whereas the slab component is spread over a range of plausible values (\emph{e.g.} Gaussian or Laplace distributions). Their product $\vy_t\cdot\Tilde{\vg}_t$ allows shrinking some values of $\Tilde{\vg}_t$ to zero and therefore effectively promotes sparsity. We further factorize the joint distribution of these variables over time as:
\begin{equation}
     p(\bar{\vg})= p(\vy_1)\prod_{t=2}^{T} p(\vy_t|\vy_{t-1})\prod_{t=1}^{T}p(\Tilde{\vg}_t)
\end{equation}
Here the conditional update $p(\vy_t|\vy_{t-1})$ is enforced to ensure that the transformation type is temporally coherent and varies sparsely. We do not enforce such constraints to $p(\Tilde{\vg}_t)$ as the Laplace distribution is very concentrated around the center and is already sparsity-inducing. 

\noindent\textbf{Spike Priors.} For the spike variable, we define the following multivariate Bernoulli prior:
\begin{equation}
\label{eq:bernoulli}
\begin{aligned}
    p(\vy_1) &= {\rm{Ber}}(P_1),\\
    p(\vy_t|\vy_{t-1}) &= {\rm{Ber}}(\sigma(a + b \vy_{t-1})).
\end{aligned}
\end{equation}
where $P_1$ is the probability of switching on, $\sigma(\cdot)$ denotes the activation function, and $a,b$ are hyper-parameters that determine the transition probability. Since we aim to obtain data sequences with smooth variations, the temporal transitions of $\vy_t$ need to be sparse. This is achieved by setting $\sigma(a)$ to be low and $\sigma(a+b)$ to be high. When drawing samples from the Bernoulli distributions in Eq.~(\ref{eq:bernoulli}), we reject all-zero samples to avoid generating sequences where no single transformations are applied. 


Fig.~\ref{fig:sequence} displays the generated sequences of MNIST~\cite{lecun1998mnist} using spike priors. The variations align with natural videos -- the transitions happen occasionally and smoothly.

\noindent\textbf{Slab Priors.} For the slab component, we use a Laplace distribution:
\begin{equation}
    p(\Tilde{\vg}_t) = {\rm{Laplace}}(\mu,\lambda) = \frac{1}{2\lambda} \exp(-\frac{|\Tilde{\vg}_t-\mu|}{\lambda})
\end{equation}
where $\mu$ is the mean, and $\lambda$ is the scale parameter that controls the sharpness of the distribution. A sharper Laplace distribution will generate speeds more peaked around $\mu$. In our experiments we set $\mu = 1$. The slab variable introduces the additional control of the transformation speed, which further mimics the dynamics of real-world videos.


\subsection{Latent Prior Time Evolution}

Based on the continuity equation $\partial_t p(\vz) = - \nabla \cdot (p(\vz) \vv(\vz))$, we can derive the probability density flow for the conditional update $p(\vz_t | \vz_{t-1}, \vg_t)$. Consider the discrete particle evolution: 
\begin{equation}
\vz_t = f(\vz_{t-1}, \vg_t) = \vz_{t-1} + \sum_k \vg_t^k \vv^k(\vz)
\end{equation}
where $\vv^k(\vz)$ denotes the velocity field of the $k$'th latent flow. We use $\vg_t$ to combine the vector fields linearly to model the possible multiple transformations. The conditional update can be derived from the change of variables formula~\cite{rezende2015variational,chen2019neural}: 
\begin{equation}
p(\vz_t | \vz_{t-1}, \vg_t)  = p(\vz_{t-1}) \Big|\frac{\mathop{d f(\vz_{t-1}, \vg_t)}}{\mathop{d\vz_{t-1}}}\Big|^{-1}
\end{equation}
In Sec.~\ref{sec:prior_evolution}, we will introduce how to define the velocity $\vv(\vz)$ such that the time evolution of $p(\vz)$ follows random trajectories as minimally informative priors.

\section{Helmholtz Flow Variational Autoencoders}


In this section, we first introduce the Helmholtz decomposition of the latent flow fields, then proceed to explain the inference over observed variables and the Optimal Transport (OT) property achieved by our posterior flow. Finally, we detail the time evolution of our latent prior and posterior.



\subsection{Helmholtz Decomposed Latent Flows}
\label{sec:helmholtz}
By the Helmholtz decomposition~\cite{helmholtz1858integrale,helmholtz1867lxiii,abraham2012manifolds}, a vector field $\mathbf{F}$ can be uniquely represented by the sum of two vector fields such that:
\begin{equation}
    \begin{aligned}
        \mathbf{F}(\mathbf{x}) & = \mathbf{G}(\mathbf{x}) + \mathbf{R}(\mathbf{x}) \\ 
    \mathbf{G}(\mathbf{x}) & = -\nabla \Phi (\mathbf{x}), \ \ \  \nabla \cdot \mathbf{R} (\mathbf{x}) = \mathbf{0}
    \end{aligned}
\end{equation}
where $\mathbf{G}(\mathbf{x})$ is the irrotational (curl-free) component ($\nabla\times\mathbf{G}(x)=0$), and $\mathbf{R}(\mathbf{x})$ is the divergence-free component. 
We then model the latent evolution using $\mathbf{F}$ as:
\begin{equation}
\begin{aligned}
    \vz_t  &= \vz_{t-1} + \sum_k \vg_t^k \mathbf{F}^k(\vz)\\
    &= \vz_{t-1} + \sum_k \Tilde{\vg}_t^k\vy_t^k\big(\nabla u^k(\vz,t) +  \vr^k(\vz)\big)
\end{aligned}
\label{eq:sample_evolution}
\end{equation}
where $u(\vz,t)=\texttt{MLP}(\vz;t)\in\mathbb{R}^1$ parameterizes the scalar spatiotemporal potential, and $\vr(\vz)=\texttt{MLP}(\vz)\in\mathbb{R}^d$ defines the divergence-free vector field.
We achieve this divergence-free constraint by imposing the following PINN loss:
\begin{equation}
    \gL_{DIV} = \frac{1}{T}\sum_t\sum_k \Big(\vg_t^k\nabla\cdot \vr^k(\vz_t)\Big)^2
\end{equation}
Richter~\emph{et al.}~\cite{richter2022neural} proposed an approach to construct strict divergence-free vector fields. However, it requires computing the full Jacobian matrix at every step, which is memory-intensive and computationally slow. For faster computation, we use a PINN to approximate the vector field. Compared with prior work \cite{song2023flow, song2023latent} which only includes the curl-free component $\mathbf{G}$, this parameterization allows for significantly increased flexibility in modeling periodic dynamics in the latent space. Furthermore, as will be illustrated later in Sec.~\ref{sec:exp_discussion}, we expect that our model automatically learns to segregate 
periodic and non-periodic transformations into these two components. 



\subsection{Inference}

We define the approximate posterior of the transformation variable $\vg_t$ to factorize as follows: 
\begin{equation}
\label{eq:q_g}
    q_\gamma(\bar{\vg}|\bar{\vx}) = \prod_{t=1}^{T} q(\vy_t | \vx_t,\vx_{t-1})q(\Tilde{\vg}_t | \vx_t,\vx_{t-1})
\end{equation}
Both the spike and slab variables are inferred from the neighboring images. For the latent particles, we have the following factorization of the approximate posterior:
\begin{equation}
\label{eq:q_z}
    q_\theta( \bar{\vz} | \bar{\vx}, \bar{\vg}) =  q(\vz_0 | \vx_0)  \prod_{t=1}^T q(\vz_t | \vz_{t-1}, \vg_t)
\end{equation}
In essence, given the transformation coefficient $\bar{\vg}$, our posterior only considers information from $\vx_0$ instead of the full sequence. However, as can be seen from Eq.~(\ref{eq:q_g}), each $\vg_t$ can see the variations happening between $\vx_t$ and $\vx_{t-1}$, and thus $\bar{\vg}$ contains the remaining sequence information.

We derive the lower bound to model evidence (ELBO) as:
\begin{equation}
    \begin{aligned}
        &\log p(\bar{\vx})= \E_{q_{\theta}(\bar{\vz}|\bar{\vx},\bar{\vg}),q_{\gamma}(\bar{\vg}|\bar{\vx})} \left[ \log
\frac{p(\bar{\vx},\bar{\vz},\bar{\vg})}{q(\bar{\vz},\bar{\vg}|\bar{\vx})}
\frac{q(\bar{\vz}|\bar{\vx},\bar{\vg})}{p(\bar{\vz}|\bar{\vx},\bar{\vg})}\right]\\
&\geq \E_{q_{\theta}(\bar{\vz}|\bar{\vx},\bar{\vg}),q_{\gamma}(\bar{\vg}|\bar{\vx})} \left[ \log\frac{p(\bar{\vx}|\bar{\vz},\bar{\vg})p(\bar{\vz}|\bar{\vg})}{q(\bar{\vz}|\bar{\vx},\bar{\vg})}\frac{p(\bar{\vg})}{q(\bar{\vg}|\bar{\vx})}\right]\\
&=\E_{q_{\theta}(\bar{\vz}|\b ar{\vx},\bar{\vg})}\left[ \log p(\bar{\vx}|\bar{\vz},\bar{\vg}) \right] + \E_{q_{\theta}(\bar{\vz}|\bar{\vx},\bar{\vg})}\left[\log\frac{p(\bar{\vz}|\bar{\vg})}{q(\bar{\vz}|\bar{\vx},\bar{\vg})}\right]\\ &+ \E_{q_{\gamma}(\bar{\vg}|\bar{\vx})}\left[\log\frac{p(\bar{\vg})}{q(\bar{\vg}|\bar{\vx})}\right]\\
    \end{aligned}
    \label{eq:elbo}
\end{equation}
The above ELBO can be further re-written as:
\begin{equation}
    \begin{aligned}
    &\log p(\bar{\vx} )\geq \sum_{t=0}^{T}\E_{q_{\theta}(\bar{\vz}| \bar{\vx},\bar{\vg})} \big[ \log p(\vx_{t}|\vz_{t},\vg_{t+1}) \big] \\&- \E_{q_{\theta}(\bar{\vz}| k)} \big[\mathrm{D}_{\text{KL}}\left[ q_{\theta}(\vz_0|\vx_0)||p(\vz_0)\right] \big] \\
    &-\sum_{t=1}^{T}\E_{q_{\theta}(\bar{\vz}| \bar{\vx},\bar{\vg})} \big[\mathrm{D}_{\text{KL}}\left[ q_{\theta}(\vz_{t}|\vz_{t-1}, \vg_t) || p(\vz_{t} | \vz_{t-1}, \vg_t) \right] \big]\\
    &-\E_{q_{\gamma}(\bar{\vg}| \bar{\vx})} \big[\mathrm{D}_{\text{KL}}\left[q_\gamma(\vy_1 | \vx_1,\vx_0)||p(\vy_1)\right] \big]\\
    &-\sum_{t=2}^{T}\E_{q_{\gamma}(\bar{\vg}| \bar{\vx})} \big[\mathrm{D}_{\text{KL}}\left[q_\gamma(\vy_t | \vx_t,\vx_{t-1})||p(\vy_t|\vy_{t-1})\right] \big]\\
    & -\sum_{t=1}^{T}\E_{q_{\gamma}(\bar{\vg}| \bar{\vx})} \big[\mathrm{D}_{\text{KL}}\left[ q_\gamma(\Tilde{\vg}_t | \vx_t,\vx_{t-1})||p(\Tilde{\vg}_t)\right] \big]
    \end{aligned}
\end{equation}
Compared with the objective of a traditional VAE, our model additionally involves the time evolution of the priors and posteriors. As noted in Sec.~\ref{sec:sp_sl}, we set $p(\Tilde{\vg}_t)$ to follow a Laplace distribution and impose multivariate Bernoulli distributions to $p(\vy_1)$ and $p(\vy_t|\vy_{t-1})$. The KL divergence on $q_\gamma(\vy_t | \vx_t,\vx_{t-1})$ serves as regularization to encourage the sparsity of $\vy_t$. That being said, the posterior $q_\gamma(\vy_t | \vx_t,\vx_{t-1})$ learns to model the transformations using as few vector fields as possible, which naturally disentangles the input variations into distinct flow fields. We apply the \texttt{Gumbel-Sigmoid} trick~\cite{jang2017categorical} for the re-parameterization and sampling of $q_\gamma({\vy}_t | \vx_t,\vx_{t-1})$.



\begin{table*}[tbp]
    \centering
    \caption{Equivariance error $\mathcal{E}_{k}$ and average log-likelihood $\log p(\vx_t)$ on MNIST~\cite{lecun1998mnist}.}
    \resizebox{0.88\linewidth}{!}{
    \begin{tabular}{c|c|c|c|c|c}
    \toprule
        \multirow{2}*{\textbf{Methods}} & \multirow{2}*{\textbf{Supervision?}} & \multicolumn{3}{c|}{\textbf{Equivariance Error ($\downarrow$)}} & \multirow{2}*{\textbf{Log-likelihood ($\uparrow$)}} \\
        \cline{3-5}
        & & \textbf{Scaling} & \textbf{Rotation} & \textbf{Coloring} & \\
        \midrule
        \textbf{VAE}~\cite{kingma2013auto} & No (\textcolor{green}{\xmark}) &1275.31$\pm$1.89 &1310.72$\pm$2.19 &1368.92$\pm$2.33 &-2206.17$\pm$1.83 \\
        \textbf{$\beta$-VAE}~\cite{higgins2016beta} & No (\textcolor{green}{\xmark}) &741.58$\pm$4.57 &751.32$\pm$5.22 &808.16$\pm$5.03 &-2224.67$\pm$2.35 \\
        \textbf{FactorVAE}~\cite{kim2018disentangling} & No (\textcolor{green}{\xmark}) &659.71$\pm$4.89 &632.44$\pm$5.76 & 662.18$\pm$5.26 &-2209.33$\pm$2.47 \\
        \midrule
        \textbf{SlowVAE}~\cite{klindt2021towards} & Weak (\textcolor{pink}{\checkmark}) &461.59$\pm$5.37 &447.46$\pm$5.46 &398.12$\pm$4.83 &-2197.68$\pm$2.39 \\
        \textbf{TVAE}~\cite{keller2021topographic} & Yes (\textcolor{red}{\cmark}) &505.19$\pm$2.77 &493.28$\pm$3.37 &451.25$\pm$2.76 &-2181.13$\pm$1.87 \\
        \textbf{PoFlow}~\cite{song2023latent} & Yes (\textcolor{red}{\cmark}) & 234.78$\pm$2.91 & 231.42$\pm$2.98 &240.57$\pm$2.58 & -2145.03$\pm$2.01 \\
        \textbf{LatentFlow}~\cite{song2023flow} & Yes (\textcolor{red}{\cmark}) &\textbf{185.42$\pm$2.35} &\textbf{153.54$\pm$3.10} &\textbf{158.57$\pm$2.95} &\textbf{-2112.45$\pm$1.57} \\
        \textbf{LatentFlow}~\cite{song2023flow} & Weak (\textcolor{pink}{\checkmark}) &{193.84$\pm$2.47} &{157.16$\pm$3.24}&{165.19$\pm$2.78} &{-2119.94$\pm$1.76} \\
        \midrule
        \rowcolor{gray!20} \textbf{STA} &No (\textcolor{green}{\xmark})& \textbf{281.32$\pm$4.71}  & \textbf{230.93$\pm$5.02} & \textbf{292.85$\pm$4.58} & \textbf{-2107.65$\pm$2.27}\\
    \bottomrule
    \end{tabular}
    }
    \label{tab:eq_mnist}
\end{table*}

\begin{figure*}[t]
    \centering
    \includegraphics[width=0.99\linewidth]{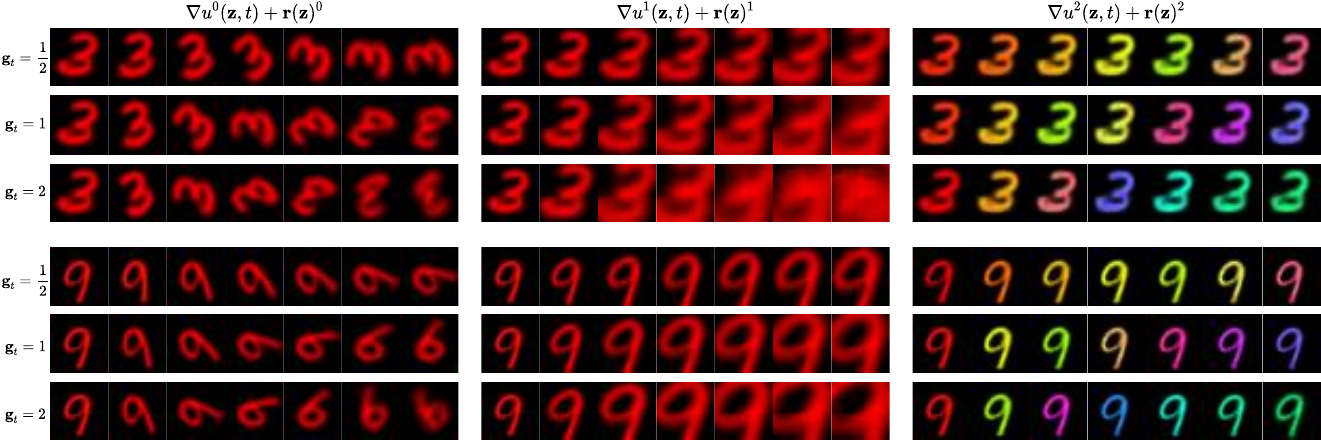}
    \caption{Traversals using individual learned flows $k{=}\{0,1,2\}$ from left to right with speeds $\vg_t{=}\{\frac{1}{2},1,2\}$ from top to bottom. }
    \label{fig:basis_flow}
\end{figure*}

\subsection{Fluid-Dynamic OT for Posterior Flow}

The divergence-free vector field is introduced to model the vorticity component of the transformation dynamics, which is particularly well-suited for capturing periodic transformations such as rotation. These vector fields are solenoidal and naturally describe closed-loop flows, making them a useful inductive bias for encoding cyclic behavior in latent dynamics. In contrast, the curl-free component corresponds to an irrotational vector field, which defines the direction of steepest descent on a scalar potential landscape. As an additional beneficial inductive bias for the path, we thus would like the potential flow to follow Optimal Transport (OT). When the vector field $\nabla u$ satisfies certain PDEs, the evolution of the probability density can be seen to minimize the $L_{2}$ Wasserstein distance between the source distribution and the target distribution. Formally, we have:

\begin{theorem}[Benamou-Brenier Formula (BBF)~\cite{benamou2000computational}]
For probability measures $\mu_{0}$ and $\mu_{1}$, the $L_{2}$ Wasserstein distance can be defined as
\begin{equation}
\begin{gathered}
    W_2(\mu_{0},\mu_{1})^2 = \min_{\rho, v} \Big\{\int\int \frac{1}{2}\rho(x,t)|v(x,t)|^2\\
    \mathop{dx}\mathop{dt}: \frac{\mathop{d}\rho(x,t)}{\mathop{dt}} = -\nabla\cdot(v(x,t)\rho(x,t))\Big\}
\end{gathered}
\end{equation}
 where the velocity $v$ satisfy: 
\begin{equation}
\begin{aligned}
    v(x,t)&=\nabla u(x,t).
\end{aligned}
\end{equation}
\end{theorem}

Solving the above equations by Karush–Kuhn–Tucker (KKT) conditions gives the optimal solution: the Hamilton-Jacobi (HJ) equation ($\partial_t u + \nicefrac{1}{2}||\nabla u||^2{=}0$). To enforce the OT property to the potential flow, we place the following PINN constraint:
\begin{equation}
\begin{gathered}
    \gL_{HJ}=\frac{1}{T}\sum_{t=1}^T\sum_k \vg_t^k\big(\frac{\partial}{\partial t} u^{k}(\vz,t) + \frac{1}{2} ||\nabla_\vz u^{k}(\vz,t)||^2\big)^2
\end{gathered}
\end{equation}
Since the linear composability $\sum_k \vg_t^k$ can be absorbed into the HJ equation, we see that our PINN loss optimizes the transportation cost for each path generated by the linear combination of $\vg_t^k$ and $\nabla u^k$.

\noindent\textbf{Assumptions and Empirical Observations.} One key assumption of the BBF is the regularity of the velocity field. Specifically, $v(x,t)$ is not only square-integrable but also typically assumed to be Lipschitz or Sobolev smooth. Although our model does not explicitly enforce such regularity, the velocity fields are defined as gradients of neural networks. In practice, we use smooth activation functions Tanh and GeLU, which make the resulting vector fields $v(z,t) = \nabla u(z,t)$ continuously differentiable and often empirically Lipschitz over compact regions of the latent space. While this does not constitute a formal guarantee, the observed smoothness provides practical support for the effectiveness of the OT-inspired Hamilton-Jacobi regularization in our framework.

\subsection{Brownian Motion for Latent Prior Evolution}
\label{sec:prior_evolution}

In line with~\cite{song2023flow}, as we do not assume any prior knowledge of each transformation, we would like to enforce minimally informative priors. This can be achieved by considering the time evolution as Brownian motion, \emph{i.e.,} random trajectories. To this end, we define the potential function $\psi^k(\vz)=-D_k\log p(\vz_t)$ which advects the density $p(\vz)$ through the induced velocity field $\nabla \psi^k(\vz)$. Then according to the continuity equation, the prior evolves as:
\begin{equation}
\begin{aligned}
    \partial_t p(\vz_t) = -\nabla\cdot\Big(p(\vz_t) v(\vz)\Big) = \sum_k(\vg_t^kD_k)\nabla^2  p(\vz_t) \\ 
\end{aligned}
\end{equation}
where $D_k$ is a learnable constant coefficient which is distinct for each $k$. The time evolution of the prior distribution thus follows a weighted diffusion process. 

\subsection{Latent Posterior Time Evolution}

We use the change of variables formula again to derive the conditional update of $q(\vz_{t}|\vz_{t-1}, \vg_t)$. Given the function of the sample evolution $\vz_{t}=h(\vz_{t-1},\vg_t)=\vz_{t-1} + \sum_k \vg_t^k\big(\nabla_{\vz} u^k + \vr^k\big)$, we still have the relation:
\begin{equation}
    q(\vz_{t}|\vz_{t-1}, \vg_t) = q(\vz_{t-1})\Big|\frac{\mathop{dh(\vz_{t-1},\vg_t)}}{\mathop{d\vz_{t-1}} }\Big|^{-1}
\end{equation}
Discretizing the continuous form and taking the logarithm yields the normalizing-flow-like density evolution:
\begin{equation}
\begin{gathered}
    \log q(\vz_{t}|\vz_{t-1}, \vg_t) = \log  q(\vz_{t-1})\\ - \log |I+\sum_k \vg_t^k ( \nabla\nabla^T u^k + \nabla(\vr^k)^T)|\\
    \approx  \log  q(\vz_{t-1}) - \sum_k \vg_t^k ( \nabla^2 u^k + \nabla\cdot\vr^k) ) \\
    = \log  q(\vz_{t-1}) - \sum_k \vg_t^k  \nabla^2 u^k 
\end{gathered}
\end{equation}
where we take a Taylor approximation to expand the probability update term and have $\nabla\cdot\vr^k=0$ by construction. We therefore expect the determinant of $\nabla \vr^k$ to be very small and hardly influence the density evolution. It is thus sufficient to not account for the impact of $\vr^k$ here. 


\begin{table*}[tbp]
    \centering
    \caption{Equivariance error $\mathcal{E}_{k}$ and average log-likelihood $\log p(\vx_t)$ on Shapes3D~\cite{3dshapes18}.}
    \resizebox{0.99\linewidth}{!}{
    \begin{tabular}{c|c|c|c|c|c|c}
    \toprule
        \multirow{2}*{\textbf{Methods}} & \multirow{2}*{\textbf{Supervision?}} & \multicolumn{4}{c|}{\textbf{Equivariance Error ($\downarrow$)}} & \multirow{2}*{\textbf{Log-likelihood ($\uparrow$)}} \\
        \cline{3-6}
        & & \textbf{Floor Hue} & \textbf{Wall Hue} & \textbf{Object Hue} & \textbf{Scale}&\\
        \midrule
        \textbf{VAE}~\cite{kingma2013auto} & No (\textcolor{green}{\xmark}) &6924.63$\pm$8.92 &7746.37$\pm$8.77 &4383.54$\pm$9.26 &2609.59$\pm$7.41 & -11784.69$\pm$4.87\\
        \textbf{$\beta$-VAE}~\cite{higgins2016beta} & No (\textcolor{green}{\xmark}) &2243.95$\pm$12.48 &2279.23$\pm$13.97 &2188.73$\pm$12.61 &2037.94$\pm$11.72 &-11924.83$\pm$5.64 \\
        \textbf{FactorVAE}~\cite{kim2018disentangling} & No (\textcolor{green}{\xmark}) & 1985.75$\pm$13.26 & 1876.41$\pm$11.93 &1902.83$\pm$12.27&1657.32$\pm$11.05&-11802.17$\pm$5.69 \\
        \midrule
        \textbf{SlowVAE}~\cite{klindt2021towards} & Weak (\textcolor{pink}{\checkmark}) & 1247.36$\pm$12.49 &1314.86$\pm$11.41 &1102.28$\pm$12.17 &1058.74$\pm$10.96 &-11674.89$\pm$5.74\\
        \textbf{TVAE}~\cite{keller2021topographic} & Yes (\textcolor{red}{\cmark}) & 1225.47$\pm$9.82 &1246.32$\pm$9.54 &1261.79$\pm$9.86 &1142.01$\pm$9.37 & -11475.48$\pm$5.18\\
        \textbf{PoFlow}~\cite{song2023latent} & Yes (\textcolor{red}{\cmark}) &885.46$\pm$10.37 &916.71$\pm$10.49 &912.48$\pm$9.86 &924.39$\pm$10.05 & -11335.84$\pm$4.95\\
        \textbf{LatentFlow}~\cite{song2023flow} & Yes (\textcolor{red}{\cmark}) & \textbf{613.29$\pm$8.93} &\textbf{653.45$\pm$9.48} &\textbf{605.79$\pm$8.63} &\textbf{599.71$\pm$9.34} &\textbf{-11215.42$\pm$5.71}\\
        \textbf{LatentFlow}~\cite{song2023flow} & Weak (\textcolor{pink}{\checkmark}) & {690.84$\pm$9.57} & {717.74$\pm$10.65}&{681.59$\pm$9.02} &{653.58$\pm$9.57} &{-11279.61$\pm$5.89}\\
        \midrule
        \rowcolor{gray!20} \textbf{STA} &No (\textcolor{green}{\xmark})&\textbf{1005.23$\pm$11.79} &\textbf{1171.69$\pm$13.64} &\textbf{928.10$\pm$11.58} & \textbf{894.77$\pm$10.94}  & \textbf{-11199.93$\pm$5.93} \\
    \bottomrule
    \end{tabular}
    }
    \label{tab:eq_shapes}
\end{table*}

\begin{figure*}[t]
    \centering
    \includegraphics[width=0.99\linewidth]{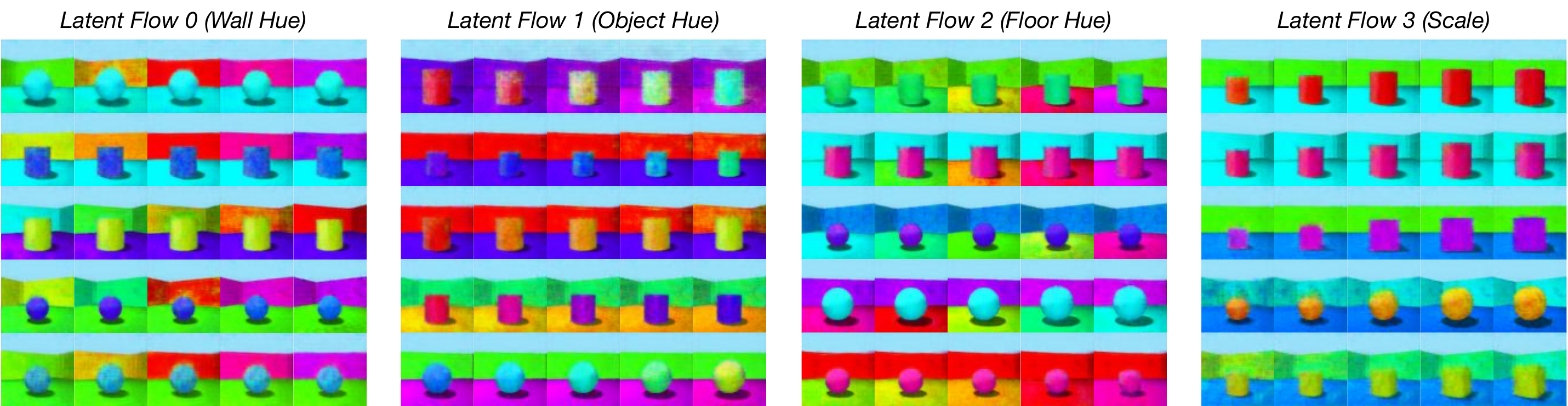}
    \caption{Traversals using each individual learned flow field on Shapes3D~\cite{3dshapes18}. In the bracket, we indicate the transformation which the traversal results look most like. Each latent flow has separate samples per row transforming from left to right.}
    \label{fig:shapes_flow}
\end{figure*}

\section{Experiments}

\subsection{Setup} 

\subsubsection{Datasets} 

We evaluate our method on two widely-used benchmarks for standard representation learning, namely MNIST~\cite{lecun1998mnist} and Shapes3D~\cite{3dshapes18}. For MNIST, The basic `pure' transformations consist of Scaling, Rotation, and Coloring. For Shapes3D, we use the self-contained four transformation primitives, including Floor Hue, Wall Hue, Object Hue, and Scale. On both datasets, we use our spike and slab prior to generate sequences that are composed of `composite' transformations. 

Beyond the toy datasets, we also evaluate our method on challenging Falcol3D and Issac3D~\cite{nie2020semi}, two complex large-scale and real-world datasets that contain sequences of different transformations. Specifically, Falcol3D consists of indoor 3D scenes with different lighting conditions and camera positions, while Isaac3D is comprised of various robot-arm movements in dynamic environments. Since the image sequences are short, we do not consider speed variations but only enforce the spike prior to generate data sequences with sparsely-varying transformations.

We further conduct some preliminary experiments of applying our method to real-world video analysis, including autonomous driving videos on Cityscape~\cite{cordts2016cityscapes} and behavior videos of social agents on CalMS~\cite{sun2021multi}. Different from the used datasets above, we directly feed raw video sequences as input and let the model discover independent motions.  

\subsubsection{Baselines} 

We compare our method with some representative approaches in the field of disentangled and equivariant representation learning, including LatentFlow~\cite{song2023flow} and PoFlow~\cite{song2023latent} which adopt potential flow to evolve the latent samples, Topographic VAE (TVAE)~\cite{keller2021topographic} which posses topographic structured latent space, SlowVAE~\cite{klindt2021towards} which proposes the sparse Laplacian prior $p(\vz_t|\vz_{t-1})=\prod \nicefrac{\alpha\lambda}{2\Gamma(1/ \alpha)}\exp{(-\lambda|z_{t,i}-z_{t-1,i}|^\alpha)}$, and $\beta$-VAE~\cite{higgins2016beta} and FactorVAE~\cite{kim2018disentangling} which encourage the factorization of the single dimensions of latent samples. We also use the vanilla VAE~\cite{kingma2013auto} as a controlled baseline. 

\subsubsection{Metrics} 

We mainly evaluate the baselines using the equivariance error which is defined as $\mathcal{E}_{k} {=} \sum_{t=1}^{T}|\vx_{t}-\texttt{Decode}(\vz_{t})|$ where $\vx_t$ is the element of sequences of each transformation primitive (\emph{e.g.,} scaling and rotation). Since our method is unsupervised, we inspect the traversal results of each basic vector field $\nabla u^k + \vr^k$ and select the index whose flow looks the most like the target transformation. The average log-likelihood of the sequence is also evaluated on the test set. Besides these two metrics, we also adopt the metric Variational Predictability (VP) score~\cite{zhu2020learning} to evaluate the disentanglement performance. Readers are kindly referred to Sec. D.3 of the supplementary for these results. 

\begin{figure*}[tbp]
    \centering
    \includegraphics[width=0.99\linewidth]{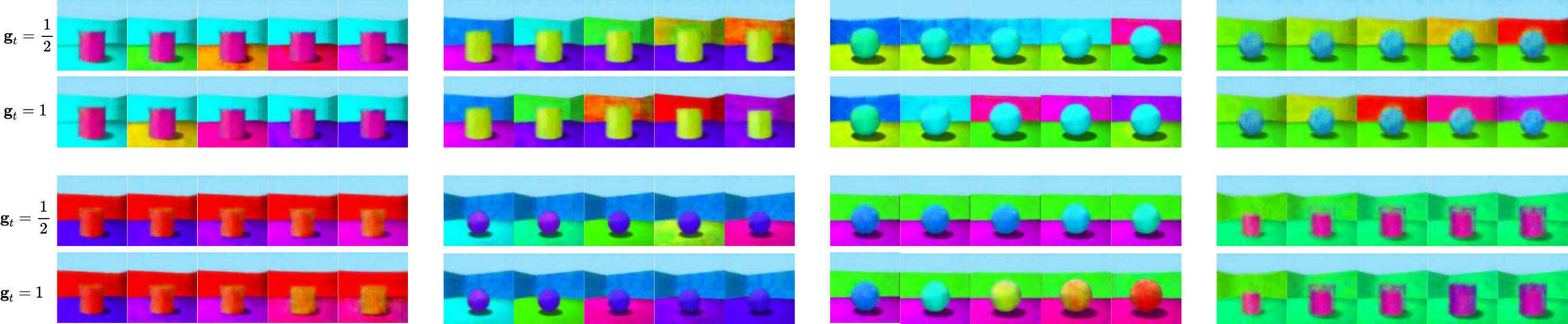}
    \caption{Traversals using learned flows with different speeds $\vg_t=\{\frac{1}{2},1\}$ on Shapes3D.}
    \label{fig:shapes_speed}
\end{figure*}

\begin{algorithm}[htbp]
\caption{Training algorithm of our method.}
\begin{algorithmic}[1]
\label{alg:training}
\REQUIRE Encoder $m$, maximum traversal step $T$, image transform function $n$, and posteriors $q_\theta$, $q_\gamma$.\\
\REPEAT
\STATE Encode: $\vz_0 = m(\vx_0)$ \\
\STATE Traversal Step Counter: $i=0$ \\
\WHILE{$i\leq T$}
\STATE Sample: $\vg_{i+1}\sim p(\vg_{i+1})$
\STATE Image transform: $\vx_{i+1} = n(\vx_i,\vg_{i+1})$
\STATE Infer: $\hat{\vg}_{i+1} = q_\gamma([\vx_i;\vx_{i+1}])$
\STATE Flow: $\vz_{i+1} = \vz_{i} + \sum \hat{\vg}_{i+1}^{k}(\nabla u^k(\vz,t)+\vr^k(\vz))$
\STATE Decode: $\vx_{i+1} = q_\theta(\vz_{i+1})$
\STATE$i=i+1$
\ENDWHILE
\STATE Optimize the ELBO $\log p(\bar{\vx})$ in Eq.~(\ref{eq:elbo}) and the PINN losses $\gL_{DIV}$ and $\gL_{HJ}$. 
\UNTIL converged
\end{algorithmic}
\end{algorithm}

\subsubsection{Implementation Details} 

Algorithm~\ref{alg:training} presents the training algorithm of our method. In practice, it is hard to learn both components ${\vy}_t$ and $\Tilde{\vg}_t$ simultaneously from the very beginning as the model could learn to use the speed $\Tilde{\vg}_t$ for choosing vector fields (by tuning the magnitude). To avoid this issue, we divide the training process into two stages. In the first stage, we only train the spike components ${\vy}_t$ to learn to select the basis vector fields. The first stage focuses exclusively on training the spike variables until convergence, which we determine based on two explicit criteria: \textbf{(1) Sparsity Criterion}: The average sparsity ratio (percentage of inactive elements) of the spike variables becomes sufficiently high (\emph{i.e,} $||\vy_t||_0 < 1+\eta$ where $\eta$ is a small tolerance constant we set to $0.3$). This ensures that the spike variables effectively select the relevant transformation vector fields. \textbf{(2) Convergence Criterion}: The approximation error stabilizes and does not further reduce, confirming that the selected vector fields adequately capture the underlying transformations. After meeting both criteria, we initiate the second training stage and introduce the slab variable $\Tilde{\vg}_t$ into the training to learn the additional control of the transformation speed. We see that this two-stage training strategy can help the optimization of these two components. We leave the rest of the implementation details to Sec. C of the supplementary material. 

\subsection{Main Results}

\subsubsection{Qualitative Results} 

Figs.~\ref{fig:basis_flow} displays the traversal results of each learned latent flow under different speeds on MNIST. Our model simultaneously disentangles the transformation categories and speeds into these vector fields in an unsupervised manner. We see that each flow field corresponds to a distinct transformation and further presents a precise control of the transformation speed. When increasing the magnitude of $\vg^k$, the transformation process will be accelerated, \emph{i.e.,} the object will rotate more degrees, get scaled with a larger factor, and change the hue more. Fig.~\ref{fig:shapes_flow} and~\ref{fig:shapes_speed} present the traversal results and the speed variations on Shapes3D. Our method still allows for disentanglement of the transformation categories and speed. We note that speed control is a major merit of our approach as the explicit control of transformation speed is seldom explored in deep representation learning.

\begin{table}[htbp]
\caption{Equivariance error $\mathcal{E}_{k}$ of composite transformations. For both baselines, we linearly combine their latent flows.}
    \centering
    \resizebox{0.99\linewidth}{!}{
    \begin{tabular}{c|c|c|c}
    \toprule
         \textbf{Methods} & \textbf{Scaling + Rotation} & \textbf{Scaling + Coloring} & \textbf{Rotation + Coloring} \\
    \midrule
         \textbf{PoFlow} &582.17$\pm$4.33 & 597.20$\pm$3.94 & 574.86$\pm$4.07 \\
         \textbf{LatentFlow}& 493.75$\pm$3.62 & 501.82$\pm$4.07 & 452.63$\pm$3.29\\
         \rowcolor{gray!20} \textbf{STA} & \textbf{293.45$\pm$4.12} &\textbf{321.82$\pm$4.74} &\textbf{407.95$\pm$4.58} \\
    \bottomrule
    \end{tabular}
    }
    \label{tab:composite}
\end{table}

\begin{table*}[htbp]
\caption{Classification accuracy (\%) of the predicted spike variable $\vy_t$ and the mean absolute error (MAE) of the slab variable $\Tilde{\vg}_t$ for different transformations on MNIST.}
    \centering
     \resizebox{0.99\linewidth}{!}{
    \begin{tabular}{c|c|c|c|c|c|c|c}
    \toprule
         Transfomration&Scaling&Rotation&Coloring&Scaling + Rotation&Scaling + Coloring&Rotation + Coloring&Scaling + Rotation + Coloring  \\
    \midrule
         Acc. ($\vy_t$) &88.74&97.46&92.85&85.49&85.32&83.14&81.94\\
         MAE ($\Tilde{\vg}_t$)&0.23&0.09&0.15&0.22&0.25&0.28&0.31\\
    \bottomrule
    \end{tabular}
    }
    \label{tab:spike_acc}
\end{table*}

\subsubsection{Quantitative Results} 

Table~\ref{tab:eq_mnist} and~\ref{tab:eq_shapes} present the evaluation results of the equivariance error and log-likelihood on MNIST and Shapes3D, respectively. We see STA achieves very competitive performance against other baselines. Specifically, STA outperforms all the unsupervised approaches by a large margin on equivariance error and rivals PoFlow~\cite{song2023latent} which requires supervision of each transformation primitive. Moreover, our method yields the highest log-likelihood on the test set, which is likely accounted for by the fact that our method incorporates a sophisticated transformation-centric prior over latent states which matches the statistics of the data. The sparse combination of multiple transformations thus can be seen as a kind of data augmentation. Among the transformations, the rotation has the smallest equivariance error. We expect this gain is largely due to the rotational vector field $\vr(\vz)$ introduced by the Helmholtz decomposition. 

Table~\ref{tab:spike_acc} compares the classification accuracy and the mean absolute error of the spike and slab variables on MNIST, respectively. Both the spike and slab variables of our STA have reasonable estimation accuracy and can recover the ground truth well. When multiple transformations are applied, we observe a slight deterioration in the results due to the increased complexity of transformation sequences. Nonetheless, the overall accuracy and error of the spike and slab components still manifest at an acceptable level.

\subsection{Discussion}
\label{sec:exp_discussion}

\begin{figure}[htbp]
    \centering
    \includegraphics[width=0.99\linewidth]{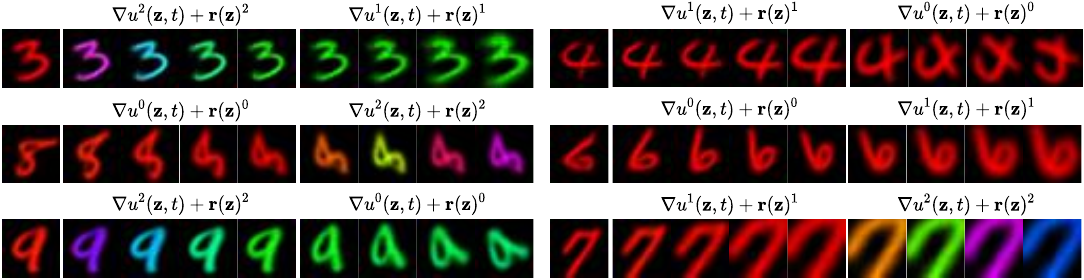}
    \caption{Traversal results of switching latent flows.}
    \label{fig:switch_flow}
\end{figure}

\subsubsection{Results on Composite Transformations} 

Besides the standard evaluation of individual transformations, it would be interesting to validate the equivariance property of composite transformations. To this end, we measure their equivariance error using the predicted spike and slab components $\vg_t$ to combine different flow fields linearly. Table~\ref{tab:composite} compares the performance against two strong baselines. Since we explicitly superpose latent flows in the training, our unsupervised STA outperforms these supervised approaches significantly, which further demonstrates the flexible linear composability of our latent flows.

\begin{figure}[htbp]
    \centering
    \includegraphics[width=0.99\linewidth]{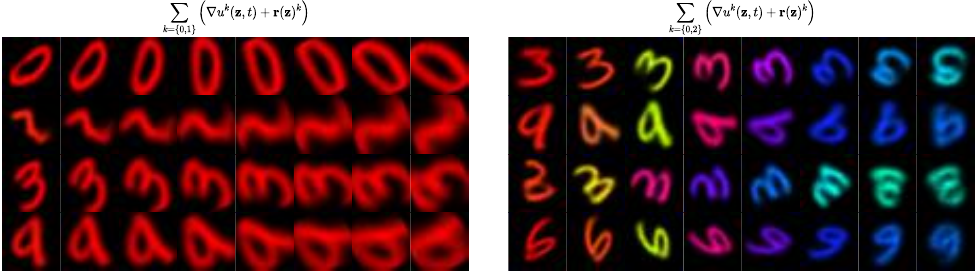}
    \caption{Traversal results of combining latent flows.}
    \label{fig:combine_flow}
\end{figure}

\subsubsection{Switchability and Composability} 

Fig.~\ref{fig:switch_flow} and~\ref{fig:combine_flow} display the traversal results of switching and combining different latent flows, respectively. Our model is able to switch to another vector field primitive with smooth output transitions and also supports performing multiple transformations simultaneously. This result indicates that our STA allows for flexible generalization to switchability and linear composability of arbitrary latent flows.

\begin{figure}[htbp]
    \centering
    \includegraphics[width=0.99\linewidth]{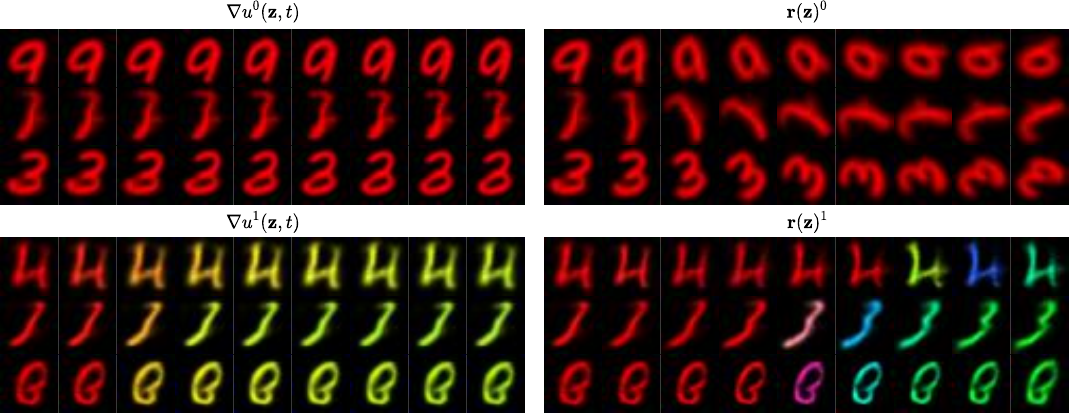}
    \caption{Traversal results using different types of vector fields.}
    \label{fig:sym_inv}
\end{figure}


\subsubsection{Periodic Transformations} 
\label{sec:period_trans}
Fig.~\ref{fig:sym_inv} compares the traversal results of two latent flows using different types of vector fields. For rotation, the divergence-free vector field $\vr^0$ dominates this transformation whereas the curl-free vector field $\nabla u^0$ has little impact. This meets our expectation that periodic transformations should be learned by rotational flow fields. For coloring, both vector fields are important and contribute to different parts of the transformations. This observation also intuitively makes sense as non-periodic transformations can be learned by both types of vector fields. Interestingly, $\nabla u^1$ mainly manipulates the image in the initial steps while $\vr^1$ takes care of the later stage, which implies that the two flow fields can complement each other in different traversal phases. 

\begin{figure*}[htbp]
    \centering
    \includegraphics[width=0.99\linewidth]{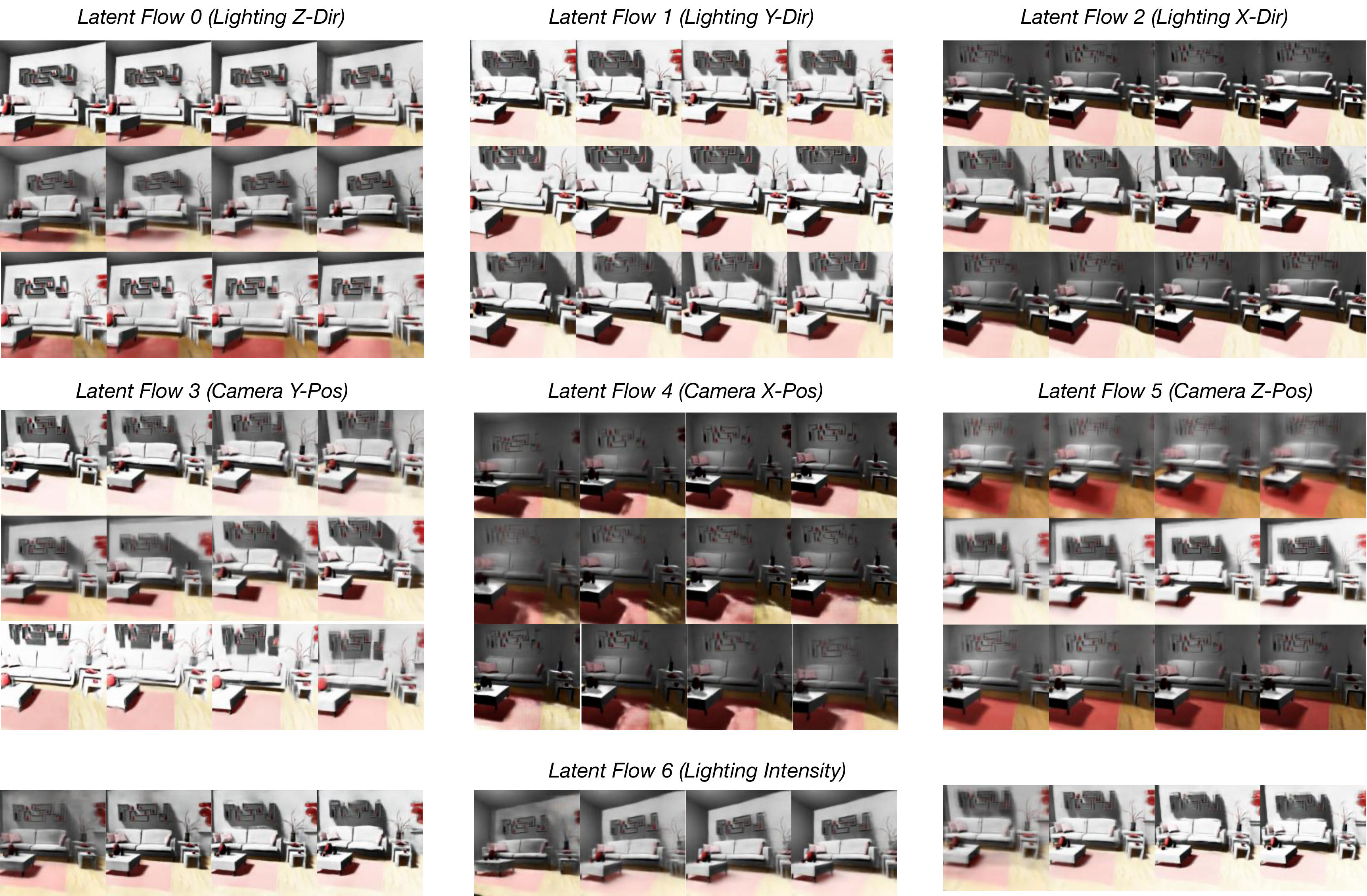}
    \caption{Traversals using each individual learned flow field on Falcol3D~\cite{nie2020semi}. In the bracket, we indicate the transformation which the traversal results look most like. Each latent flow has separate samples per row transforming from left to right. The bottom row displays the traversal result generated by the $6$'th latent flow field.}
    \label{fig:falor_flow}
\end{figure*}

\begin{table*}[htbp]
    \centering
    \caption{Equivariance error $\mathcal{E}_{k}$ on Falcor3D.}
     \resizebox{0.99\linewidth}{!}{
    \begin{tabular}{c|c|c|c|c|c|c|c}
    \toprule
         \textbf{Methods}&	\textbf{Lighting Intensity}	&\textbf{Lighting X-dir}	&\textbf{Lighting Y-dir}	&\textbf{Lighting Z-dir}&	\textbf{Camera X-pos}&	\textbf{Camera Y-pos}&	\textbf{Camera Z-pos}  \\
        \bottomrule
        \textbf{TVAE}~\cite{keller2021topographic}	&11477.81	&12568.32	&11807.34	&11829.33	&11539.69	&11736.78	&11951.45 \\
        \textbf{PoFlow}~\cite{song2023latent}	&8312.97	&7956.18&	8519.39	&8871.62	&8116.82	&8534.91	&8994.63 \\
        \textbf{LatentFlow}~\cite{song2023flow}&	5798.42	&6145.09	&6334.87	&6782.84	&6312.95&	6513.68&	6614.27\\
        \bottomrule
         \rowcolor{gray!20}\textbf{STA}&	8672.91	&8146.91&	8729.06	&9023.56	&8064.75	&8856.92&	9134.02 \\
    \bottomrule
    \end{tabular}
    }
    \label{tab:eq_falcol}
\end{table*}


\subsection{Learning Separate Controls of Vector Fields}
\label{sec:sep_control}

With a slight modification to our method, each transformation primitive can be associated with a specific vector field, which could make the Helmholtz decomposition more compelling. To this end, we can introduce separate controls ${\vy_1}_t,{\vy_2}_t$ for the curl-free and divergence-free vector fields:
\begin{equation}
    \vz_t = \vz_{t-1} + \sum_k \Tilde{\vg}_t^k \Big({\vy_1}_t^k\nabla u^k(\vz,t)+{\vy_2}_t^k \vr^k(\vz)\Big)
\end{equation}
The above formulation slightly modifies Eq.~\ref{eq:sample_evolution} in controlling the sample evolution. The two vector fields therefore share the same speeds while having separate switches. This increases the flexibility of choosing flow fields, thus matching the goal of learning to segregate the symmetries and invariances. For the posterior, we use the analytical representation of the OR gate to compose $\mathbf{y}_t$ as:
\begin{equation}
    \vy_t = {\vy_1}_t + {\vy_2}_t - {\vy_1}_t {\vy_2}_t
\end{equation}
This means that if either ${\vy_1}_t$ or ${\vy_2}_t$ is active, their 'global' spike variable $\vy_t$ will be active. Accordingly, the posterior $q_\gamma(\vy_t | \vx_t,\vx_{t-1})$ is changed to $q_\gamma({\vy_1}_t,{\vy_2}_t | \vx_t,\vx_{t-1})$ to allow for inferring controls of the decomposed vector fields. As for the priors, we simply sample ${\vy_1}_t, {\vy_2}_t$ from the candidates $\{10,01,11\}$ if $\vy_t$ is active. 

\begin{table}[htbp]
    \centering
    \caption{The learned association of different vector fields for each transformation on MNIST.}
    \begin{tabular}{c|c|c|c}
     \toprule
         Seed & Scaling & Rotation & Coloring  \\
    \midrule
          42 & $\nabla u^0(\vz)$ & $\vr^1(\vz)$& $\vr^2(\vz)$ \\
          3857 & $\nabla u^0(\vz)$ + $\vr^0(vz)$ & $\vr^1(\vz)$& $\vr^2(\vz)$\\
    \bottomrule
    \end{tabular}
    \label{tab:sep_control}
\end{table}

Table~\ref{tab:sep_control} displays the vector field correspondences using separate controls with different random seeds. For periodic transformations like rotation, our model learns to associate the flow with a divergence-free vector field. In contrast, the non-periodic transformations are modeled either by a curl-free field alone or by the combination of both flow fields. The results are very coherent with the analysis in Sec.~\ref{sec:period_trans} that the two vector fields play different roles in modeling transformations. Further, the separate control justifies the application of the Helmholtz decomposition in learning latent flows for flexibly modeling input transformations. We do not present the decomposed controls as the main approach because the training can be non-trivial if we further introduce slab variables for speed variations. Nonetheless, we empirically find that this approach works well when there is only the spike component to be modeled. 

\subsection{Real-world Video Analysis}
\label{sec:real_video}

\begin{figure*}[tbp]
    \centering
    \includegraphics[width=0.99\linewidth]{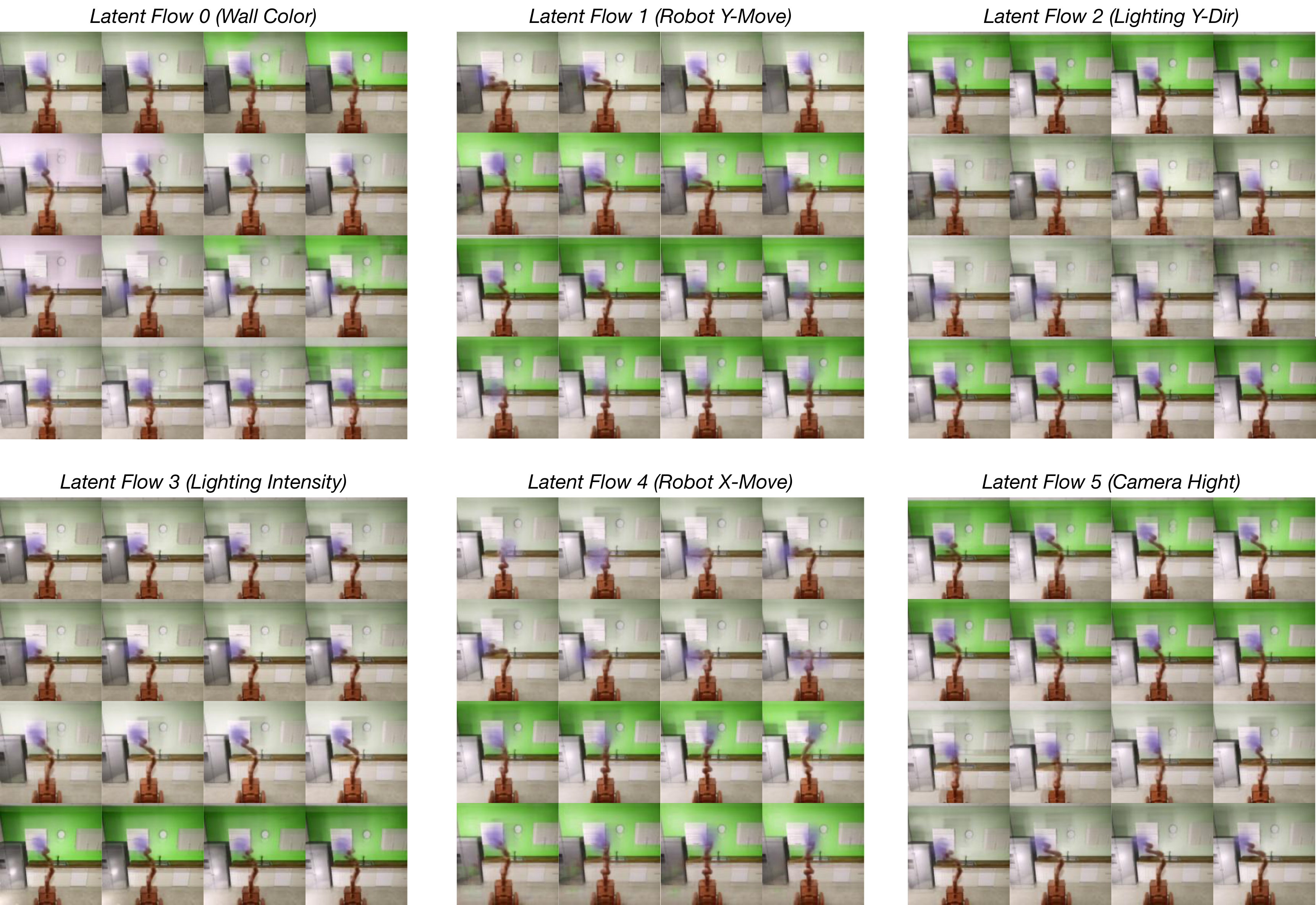}
    \caption{Traversals using each individual learned flow field on Issac3D~\cite{nie2020semi}. In the bracket, we indicate the transformation which the traversal results look most like. Each latent flow has separate samples per row transforming from left to right.}
    \label{fig:issac_flow}
\end{figure*}

\begin{table*}[htbp]
    \centering
    \caption{Equivariance error $\mathcal{E}_{k}$ on Issac3D.}
     \resizebox{0.99\linewidth}{!}{
    \begin{tabular}{c|c|c|c|c|c|c|c|c}
    \toprule
         Methods	&Robot X-move	&Robot Y-move	&Camera Height	&Object Scale&	Lighting Intensity&	Lighting Y-dir&	Object Color	&Wall Color \\
        \bottomrule
        \textbf{TVAE}~\cite{keller2021topographic}	&8441.65	&8348.23	&8495.31	&8251.34	&8291.70&	8741.07	&8456.78	&8512.09\\
        \textbf{PoFlow}~\cite{song2023latent}	&6572.19	&6489.35	&6319.82	&6188.59	&6517.40&	6712.06&	7056.98	&6343.76\\
        \textbf{LatentFlow}~\cite{song2023flow}&	3659.72&	3993.33&	4170.27&	4359.78&	4225.34&	4019.84&	5514.97&	3876.01\\
        \bottomrule
         \rowcolor{gray!20}\textbf{STA}&	7012.34&	6399.57&	6589.48&	6104.74&	6298.16&	6517.23&	6674.98&	6519.38\\
    \bottomrule
    \end{tabular}
    }
    \label{tab:eq_issac}
\end{table*}

\subsubsection{Robot Arms and Indoor Scenes} 

Fig.~\ref{fig:falor_flow} and~\ref{fig:issac_flow} show the learned latent flows on Falcol3D and Issac3D~\cite{nie2020semi}, respectively. As can be seen from the figures, even on these challenging large-scale datasets, our method still allows for unsupervised disentanglement of complex real-world transformations. Table~\ref{tab:eq_falcol} and~\ref{tab:eq_issac} compares the equivariance error on the two datasets. Similar to the results on MNIST and Shapes3D, our method achieves very competitive performance against supervised ones. This demonstrates that the proposed sparsity priors also scale up to sequences of complex transformations.

\begin{figure}[htbp]
    \centering
    \includegraphics[width=0.99\linewidth]{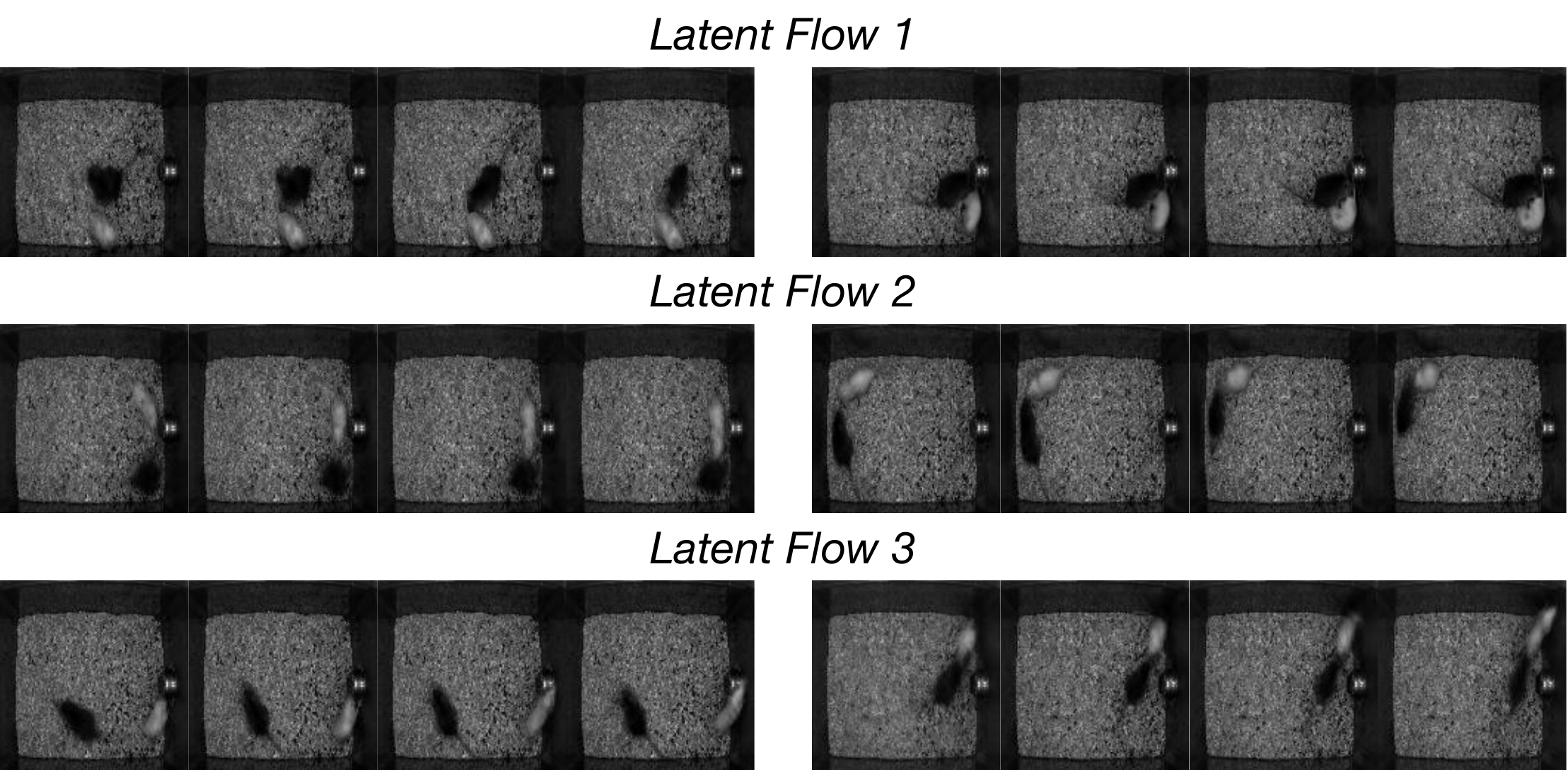}
    \caption{Traversal results of learned latent flows on CalMS~\cite{sun2021multi}. For each latent flow, we display two exemplary sequences, and the flow transforms the image from left to right.}
    \label{fig:flow_calms}
\end{figure}

\subsubsection{Agent Behavioral Videos}

\begin{table}[htbp]
    \centering
    \caption{Behavior classification results on CalMS~\cite{sun2021multi}.}
    \resizebox{0.99\linewidth}{!}{
    \begin{tabular}{c|c|c|c|c}
     \toprule
         Method  & MARS~\cite{segalin2021mouse} &B-Kind~\cite{sun2021multi} & Trajectory-LSTM~\cite{sun2021multi} & \cellcolor{gray!20} STA \\
            \midrule
         Supervision? & Yes (\textcolor{red}{\cmark}) & Yes (\textcolor{red}{\cmark})& Yes (\textcolor{red}{\cmark}) & \cellcolor{gray!20} No (\textcolor{green}{\xmark})\\
         mAP & 0.880 & 0.852 & 0.712 &\cellcolor{gray!20} 0.793\\
    \bottomrule
    \end{tabular}
    }
    \label{tab:calms_class}
\end{table}

\begin{figure*}[t]
    \centering
    \includegraphics[width=0.99\linewidth]{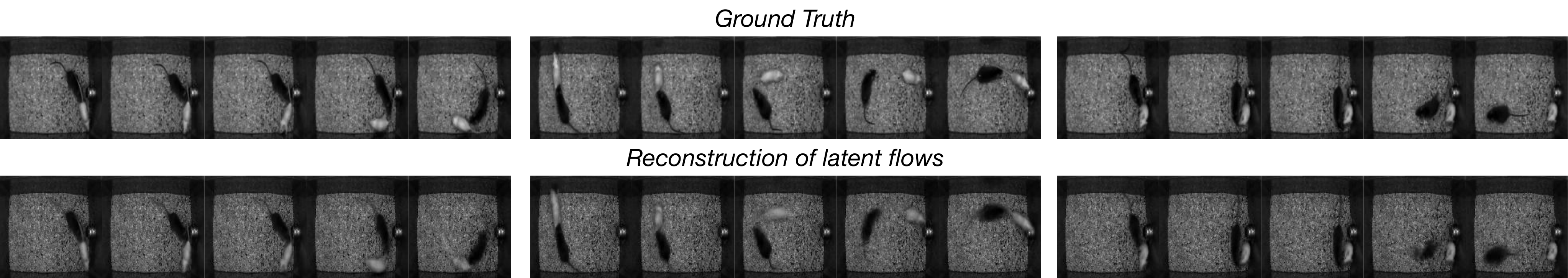}
    \caption{Exemplary comparisons of the ground truth image sequences and reconstruction results. For each sequence, we start with reconstructing the initial frame and use the spike component and latent flow fields to generate the rest frames.}
    \label{fig:calms_recon}
\end{figure*}

\begin{figure}[t]
    \centering
    \includegraphics[width=0.99\linewidth]{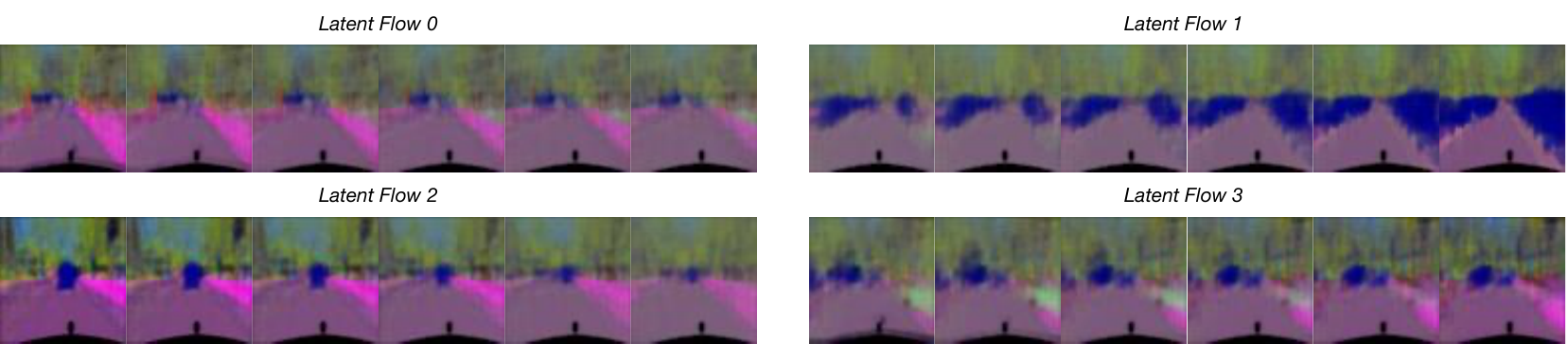}
    \caption{Traversals results of learned flow fields on downsampled segmentation masks of Cityscape~\cite{cordts2016cityscapes}. Each latent flow transforms the image from left to right.}
    \label{fig:cityscape_flow}
\end{figure}

We apply our method to disentangle the complex social interactions of mice on  Task $1$ of CalMS~\cite{sun2021multi}. On this dataset, there exist three ground truth interactions, namely `investigation', `attack', and `mount'. We thus define $3$ latent flows and let the model learn the interactions from the raw videos. The images are of the resolution $128\times128$. 

Fig.~\ref{fig:flow_calms} displays the exemplary traversal results of three distinct latent flows. We could have reasonable interpretations of the interaction categories for these latent flows. Specifically, we might interpret latent flow $1$ as `investigation', latent flow $2$ as `attack', and latent flow $3$ as `mount', respectively. To validate if the interpretations align with human annotations, we compute the correlation between the predicted spike variable and the behavior labels. Table~\ref{tab:calms_class} reports the classification accuracy of each interaction class. As an unsupervised approach, our method achieves competitive results against the supervised baselines, indicating that the sparsity prior can help disentangle the mouse behaviors. Fig.~\ref{fig:calms_recon} displays a few examples of image sequences and the reconstruction results using the spike prior and the latent flow fields. Our STA can reconstruct the behaviors that are close to the ground truth.



\subsubsection{Autonomous Driving Videos}  

Finally, we take a step further to evaluate our method on Cityscape~\cite{cordts2016cityscapes}, the challenging real-world autonomous driving videos. We take the sequences of segmentation masks as the training data and downsample the resolution to $64{\times}64$. Fig.~\ref{fig:cityscape_flow} displays some exemplary traversals of different latent flows. On this dataset, there are no ground truth generative factors so we may have some reasonable interpretations according to the disentangled transformations: we may interpret latent flow $0$ as turning left (the sidewalk region on the right side shrinks), latent flow $1$ as getting closer to the front car (the car region expands), latent flow $2$ as getting away from the front car (the car region shrinks and disappears), and latent flow $3$ as changing the right side from terrain to sidewalk. \textit{Notice that this is an initial attempt to apply our method to complex real-world video analysis.} Nonetheless, this preliminary experiment demonstrates that our method could have real-world applicability for video understanding. 

\section{Conclusion}

Inspired by the sparsity in natural data statistics, we propose a new generative modeling framework which model composite input transformations as sparse combinations of learned vector fields. We leverage the Helmholtz decomposition to parameterize flexible latent flows, and the sparse combination is further learned as a latent variable following the spike and slab prior. We train our model using the standard variational objective entirely unsupervised. Extensive experiments demonstrate that our model yields the state of the art in unsupervised approximate equivariance and archives the highest likelihood in modeling sequences. Our method can segregate periodic and non-periodic transformations and supports flexibly switching or combining latent flows. Our framework also allows for controlling the transformation speed by tuning the stepsizes of the latent flows. We expect our STA to pave the way for more research in unsupervised representation learning for approximate equivariance.

\section*{Acknowledgments}

This research was supported in part by gifts from Cisco and OpenAI. 

%
\IEEEpeerreviewmaketitle


%


\ifCLASSOPTIONcaptionsoff
  \newpage
\fi



%
\bibliographystyle{IEEEtran}
\bibliography{egbib}
%

%
\begin{IEEEbiography}[{\includegraphics[width=1in,height=1.25in,keepaspectratio]{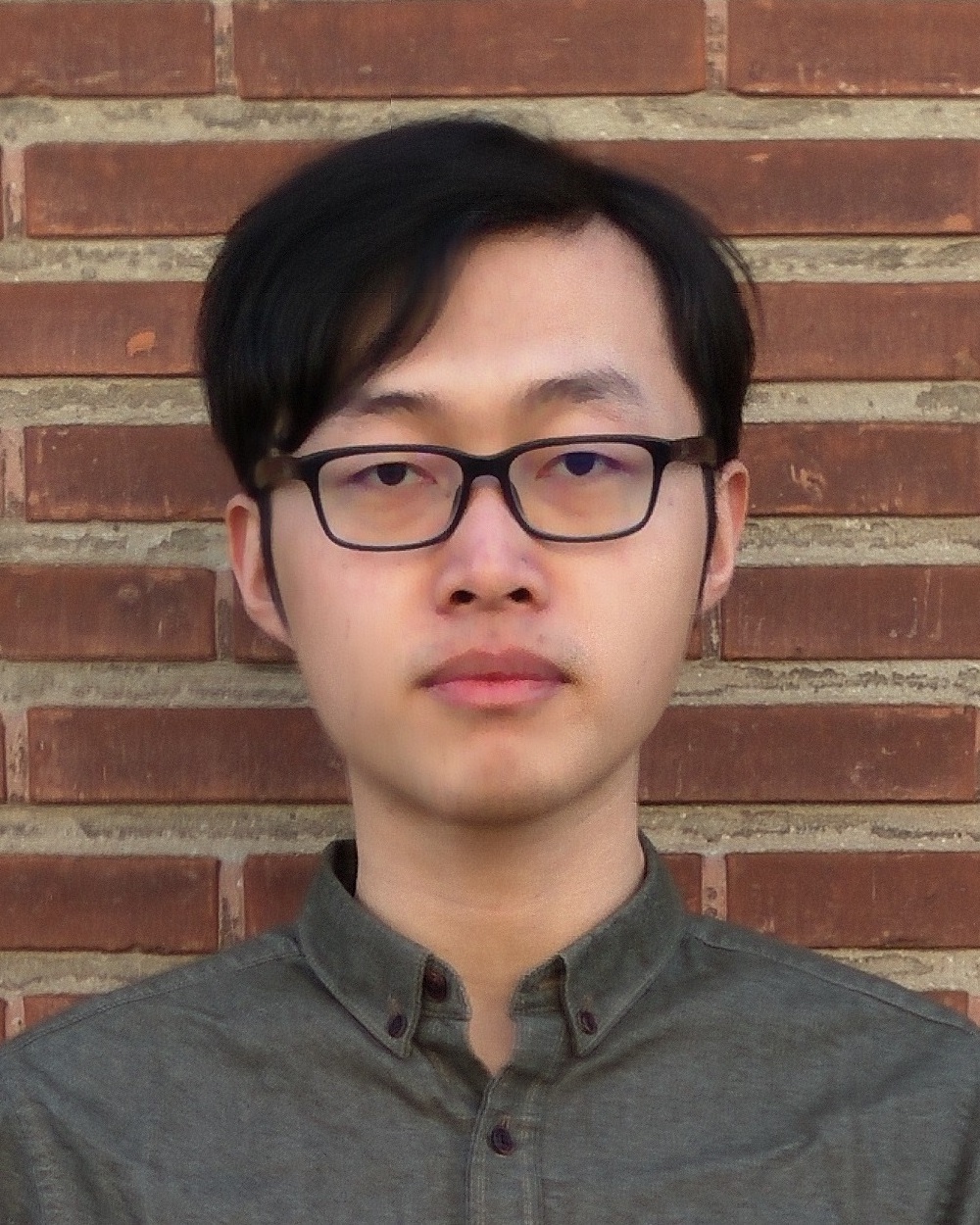}}]{Yue Song}
received the B.Sc. \emph{cum laude} from KU Leuven, Belgium and the joint M.Sc. \emph{summa cum laude} from the University of Trento, Italy and KTH Royal Institute of Technology, Sweden, and the Ph.D. \emph{summa cum laude} from the Multimedia and Human Understanding Group (MHUG) at the University of Trento, Italy. Currently, he is a post-doctoral research associate at Caltech. His research interests are structured representation learning.
\end{IEEEbiography}

\begin{IEEEbiography}[{\includegraphics[width=1in,height=1.25in,keepaspectratio]{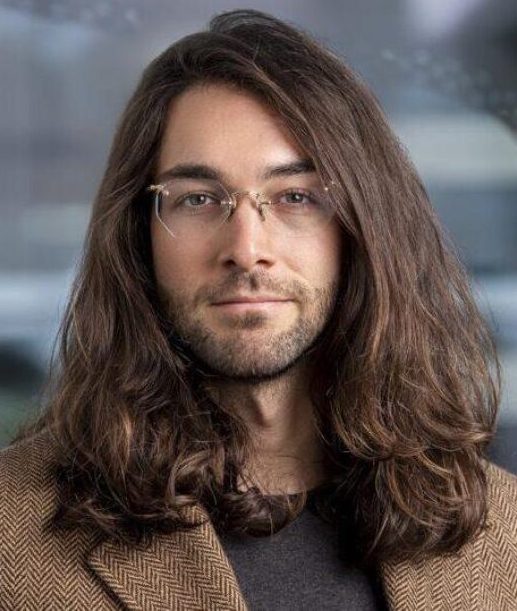}}]{T. Anderson Keller} is a postdoctoral research fellow at the Kempner Institute at Harvard University. He completed his doctorate under the supervision of Max Welling at the University of Amsterdam. His current research focuses on structured representation learning, probabilistic generative modeling, and biologically plausible learning. His research explores ways to develop deep probabilistic generative models that are meaningfully structured with respect to observed, real-world transformations.
\end{IEEEbiography}


\begin{IEEEbiography}
[{\includegraphics[width=1in,height=1.25in,clip,keepaspectratio]{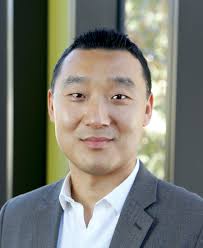}}]
{Yisong Yue} is a professor of Computing and Mathematical Sciences at Caltech. My research interests lie primarily in machine learning, and span the entire theory-to-application spectrum from foundational advances all the way to deployment in real systems. I work closely with domain experts to understand the frontier challenges in applied machine learning, distill those challenges into mathematically precise formulations, and develop novel methods to tackle them.
\end{IEEEbiography}

\begin{IEEEbiography}
[{\includegraphics[width=1in,height=1.25in,clip,keepaspectratio]{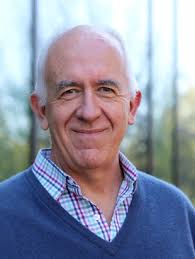}}]
{Pietro Perona} received the graduate degree in electrical engineering from the Università di Padova in 1985 and the PhD degree in electrical engineering and computer science from the University of California at Berkeley in 1990. After a postdoctoral fellowship at MIT in 1990-1991 he joined the faculty of the California Institute of Technology, Caltech in 1991, where he is now an Allen E. Puckett professor of electrical engineering and computation and neural systems. His current interests include visual recognition, modeling vision in biological systems, modeling and measuring behavior, and Visipedia. He has worked on anisotropic diffusion, multiresolution-multiorientation filtering, human texture perception and segmentation, dynamic vision, grouping, analysis of human motion, recognition of object categories, and modeling visual search.
\end{IEEEbiography}

\begin{IEEEbiography}
[{\includegraphics[width=1in,height=1.25in,clip,keepaspectratio]{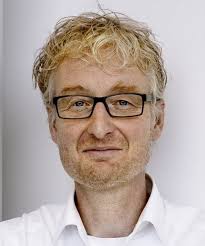}}]
{Max Welling} is a research chair in Machine Learning at the University of Amsterdam and a Distinguished Scientist at MSR. He is a fellow at the Canadian Institute for Advanced Research (CIFAR) and the European Lab for Learning and Intelligent Systems (ELLIS) where he also serves on the founding board. His previous appointments include VP at Qualcomm Technologies, professor at UC Irvine, postdoc at U. Toronto, and UCL under the supervision of Prof. Geoffrey Hinton, and postdoc at Caltech under the supervision of Prof. Pietro Perona. He finished his PhD in theoretical high energy physics under the supervision of Nobel laureate Prof. Gerard ‘t Hooft. 
\end{IEEEbiography}
\newpage

\appendices
\newpage
\section{Limitations and Future Work} 

\noindent\textbf{Limitations.} One of the main limitations of this work in relation to prior work on independent component analysis and sparse coding is the unanswered question of `identifiability' of latent factors in our model. Specifically, much of the early work on ICA was focused around answering to what extent the ground truth latent variables can be recovered after an unknown linear transformation \cite{comon1994independent, 1306473}. More recently, a number of papers have demonstrated the necessary conditions for these identifiability proofs to extend to non-linear transformations of the latent variables \cite{hyvarinen2023nonlinear}. Interestingly, these necessary conditions often revolve around temporal structure such as time-step conditioning \cite{pmlr-v89-hyvarinen19a}, or sparsity of transformations over time \cite{klindt2021towards}. Our model, Sparse Transformation Analysis, clearly takes inspiration from these proofs by integrating these factors into a highly flexible generative modeling framework. Empirically, we demonstrate that STA does separate the ground-truth factors of variation from sequence data entirely unsupervised for a variety of datasets, including `real world' datasets with more natural image statistics. We note that even in the case of models which have proved this identifiability, this is often only under strict assumptions about the true data distribution, and these assumptions are frequently seen to be invalidated on real data. These papers therefore similarly resort to measuring if their models are able to separate the underlying factors of variations in an empirical manner, as we have done in this paper. Furthermore, in our empirical analysis, we see that due to the flexibility endowed by the learned latent flows of our model, it is able to outperform models which indeed have these identifiability guarantees (such as \cite{klindt2021towards}). That being said, in future work, we believe that it would undoubtedly be beneficial and of great interest to understand the exact necessary conditions for such a model to provably identify the generative factors. We believe such an understanding will likely be helpful for selecting optimal settings of the hyperparameters, such as the probability of switching on for the Bernoulli distribution or the scale parameter for the Laplace distribution, while similarly helping to guide the future development of unsupervised representation learning in general.

\noindent\textbf{Generating High-Resolution Videos.} When dealing with high-resolution videos, the complexity of detailed objects and diverse backgrounds substantially increases, making it extremely challenging to effectively capture intricate motions or transformations in the highly compressed latent space. To address these challenges, incorporating additional sophisticated feature extraction methods (such as key-point tracking) could be beneficial, as demonstrated in previous motion tracking or synthesis work~\cite{sun2021multi,tulyakov2018mocogan}. Thus, the primary obstacles lie in designing powerful motion feature extraction methods and identifying meaningful latent representations that can robustly encode complex motion dynamics at higher resolutions into lower dimensions.

\noindent\textbf{Integration into Diffusion Models.} Another critical obstacle is that standard diffusion models do not inherently define a semantically meaningful, highly compressed latent space. To integrate our methodology into diffusion-based models, we must first carefully identify an appropriate latent representation. One potential approach is leveraging the ``h-space" proposed by Kwon~\emph{et al.}~\cite{kwon2022diffusion}, which uses the bottleneck features from the score-prediction U-Net as a latent representation. However, empirical validation is still necessary to confirm whether such latent spaces are sufficiently powerful to capture transformations effectively compared to latent spaces explicitly constructed by GANs or VAEs.

\section{Identifiability of Sparse Transformation Analysis}

We provide a formal argument supporting the identifiability of our sparse transformation model under mild assumptions. Specifically, we show that the latent vector fields and sparse transformation coefficients are identifiable (up to permutation and scaling) given observed sequential data.

\subsection{Assumptions}

Let $\vx_t \in \mathbb{R}^D$ be the observation at time $t$, and $\vz_t \in \mathbb{R}^d$ be the latent code. The latent evolution follows:
\begin{equation}
\vz_t = \vz_{t-1} + \sum_{k=1}^{K} \vg_t^k \vv_k(\vz_{t-1}),
\end{equation}
where $\vv_k(z) = \nabla \vu_k(z) + \vr_k(z)$ is the $k$-th vector field (Helmholtz decomposition), and $\vg_t \in \mathbb{R}^K$ is a sparse coefficient vector. We make the following assumptions:
\begin{itemize}
\item \textbf{A1 (Smooth Decoding)}: The observation mapping $f: \mathbb{R}^d \rightarrow \mathbb{R}^D$ is a diffeomorphism. Thus, $\vx_t = f(\vz_t)$ and $\vz_t$ can be recovered from $\vx_t$.
\item \textbf{A2 (Sparse Composition)}: The transformation coefficients $\vg_t$ are $s$-sparse: $||\vg_t||_0 \leq s$ with $s \ll K$.
\item \textbf{A3 (Vector Field Independence)}: For almost every $\vz$, the set $\{\vv_k(\vz)\}_{k=1}^K$ is linearly independent.
\item \textbf{A4 (Sufficient Support Diversity)}: The dataset contains a set of time steps $t$ such that the support patterns $\mS_t = \text{supp}(\vg_t)$ vary and sufficiently span all combinations of up to $s$ active components.
\end{itemize}

\subsection{Theorem (Identifiability)}

\textit{Under assumptions A1--A4, the set of vector fields $\{\vv_k(\cdot)\}_{k=1}^K$ and the sparse transformation coefficients $\{\vg_t\}_{t=1}^T$ are identifiable up to permutation and scaling.}

\subsection{Proof}
From \textbf{A1}, we can invert the observations to obtain the latent sequence $\{\vz_t\}$. Define the latent displacement:
\begin{equation}
\delta \vz_t := \vz_t - \vz_{t-1} = \sum_{k \in \mS_t} \vg_t^k \vv_k(\vz_{t-1}),
\end{equation}
where $\mS_t = \text{supp}(\vg_t)$ and $|\mS_t| \leq s$.

At each $t$, $\delta \vz_t$ is a sparse linear combination of the set $\{\vv_k(\vz_{t-1})\}$. Under A3 and A4, we obtain a sequence of sparse coding problems:
\begin{equation}
\delta \vz_t = \mV(\vz_{t-1}) \vg_t, \quad \text{with } ||\vg_t||_0 \leq s,
\end{equation}
where $\mV(\vz_{t-1}) = [\vv_1(\vz_{t-1}), \dots, \vv_K(\vz_{t-1})] \in \mathbb{R}^{d \times K}$.
From the theory of sparse dictionary learning~\cite{donoho2003optimally}, if a dictionary $\mV$ has columns that are linearly independent and the sparsity level $s$ satisfies $s < \text{spark}(\mV)/2$, then for every vector $\delta\vz$ that admits a $s$-sparse representation over $\mV$, this representation is unique. The \textit{spark} of a matrix is the smallest number of columns that are linearly dependent. If $\mV$ has full column rank and no subset of $2s$ or fewer columns is linearly dependent, then the sparse coefficients and the dictionary can be uniquely recovered (up to permutation and scaling) from sufficiently many samples.

Moreover, under the assumption A4, the dataset provides a diverse set of support patterns $\mS_t$, ensuring that the different combinations of active vector fields are sufficiently sampled. This property is crucial for the joint recovery of the dictionary $\mV(\vz)$ and sparse codes $\vg_t$ using methods akin to K-SVD~\cite{aharon2006k} or ER-SpUD~\cite{spielman2012exact}. When such diversity holds across $\vz$, the function-valued dictionary ${\vv_k(\vz)}$ can be recovered pointwise, as each $\vz_{t-1}$ provides local linear constraints.

Therefore, the latent transformation structure -- both vector fields $\{\vv_k\}$ and sparse codes $\{\vg_t\}$ -- is identifiable up to permutation and scaling.

\subsection{Identifiability of Helmholtz Components}

This identifiability result also applies to the individual components of the Helmholtz decomposition. Specifically, the divergence-free components ${\vr_k}$ (vorticity) and the curl-free components ${\nabla \vu_k}$ (potential flows) are each identifiable up to permutation and scaling, provided that they appear independently or in sufficiently varied combinations within the support patterns. Since the decomposition is additive and the dictionary learning problem is posed over the combined flows $\vv_k = \nabla \vu_k + \vr_k$, the linear independence and sparse excitation across time guarantee that both components can be separately identified as long as they do not systematically co-occur. In practice, this holds because different transformation types (\emph{e.g.,} rotation vs. scaling) are encoded by different structural priors and appear in distinct contexts.

\section{Proof of how the Hamilton-Jacobi equation solves Optimal Transport}

\begin{theorem}[Benamou-Brenier Formula~\cite{benamou2000computational}]
For probability measures $\mu_{0}$ and $\mu_{1}$, the $L_{2}$ Wasserstein distance can be defined as
\begin{equation}
\begin{gathered}
    W_2(\mu_{0},\mu_{1})^2 = \min_{\rho, v} \Big\{\int\int \frac{1}{2}\rho(x,t)|v(x,t)|^2\\
    \mathop{dx}\mathop{dt}: \frac{\mathop{d}\rho(x,t)}{\mathop{dt}} = -\nabla\cdot(v(x,t)\rho(x,t))\Big\}
\end{gathered}
\end{equation}
 where the velocity $v$ satisfy: 
\begin{equation}
\begin{aligned}
    v(x,t)&=\nabla u(x,t).
\end{aligned}
\end{equation}
\end{theorem}

We now prove why the Hamilton-Jacobi equation solves the Optimal Transport (OT) problem. Let us define the momentum $m=\rho v$ and introduce a Lagrange multiplier $u$ for the continuity equation ($\partial_t \rho = -\nabla\cdot(v\rho) =-\nabla\cdot m$). The corresponding Lagrangian function would be given by:
\begin{equation}
    \begin{aligned}
    L(\rho, m, \phi) & = \int_D \int_0^1 \frac{||m||^2}{2\rho} + u (\partial_t \rho  + \nabla\cdot m)
    \end{aligned}
\end{equation}
where the second term is the equality constraint of the weak condition. Exploiting the integration by parts formula, we can re-write the above equation as
\begin{equation}
    L(\rho, m, \phi) = \int_D \int_0^1 \frac{||m||^2}{2\rho} + \int_D u\rho |_0^1 - \int_D \int_0^1 (\partial_t u\rho + \nabla u\cdot m)
\end{equation} 
Applying the set of Karush–Kuhn–Tucker (KKT) conditions ($\partial_m L=0$, $\partial_u L=0$, and $\partial_\rho L =0$) directly gives:
\begin{equation}
    \begin{cases}
    \partial_m L = v - \nabla u = 0\\
    \partial_u L = \partial_t \rho  + \nabla\cdot m = 0\\
 \partial_\rho L = - \frac{||m||^2}{2\rho^2} - \partial_t u = -\frac{1}{2}||v||^2 - \partial_t u =0\\
    \end{cases}
\end{equation}
where the first condition indicates that the velocity field $v(x,t)$ is given by the gradient $\nabla u(x,t)$, the second condition gives the continuity equation which holds in the sense of distributions, and the third condition yields the optimal solution for minimizing the Wasserstein distance — the Hamilton-Jacobi equation ($\partial_t u + \frac{1}{2}||\nabla u||^2 = 0$).

\section{Additional Implementation Details}
\label{sec:exp_detail}

\subsection{Model Architectures}

For the $\texttt{MLPs}$ that parameterize the scalar potential $u(\vz,t)$ and the divergence-free vector field $\vr(\vz)$, we use linear layers to encode the latent samples and use $\texttt{Tanh}$ as the activation function. The sinusoidal positional embeddings~\cite{vaswani2017attention} are used to embed the timestep $t$. For our variational auto-encoders, the encoder simply takes four stacked convolution layers with ReLU activation functions, while the decoder consists of four transposed convolution layers.

\subsection{Data Sequences and Baselines}

For the spike component $\vy_t$, we set $P_1$ to $0.1$ for the initial Bernoulli prior $p(\vy_1)$ and set $\sigma(a)=0.1,\sigma(a+b)=0.9$ for the conditional update $p(\vy_t|\vy_{t-1})$. Each transformation primitive has a probability of $0.1$ to be picked in the initial timestep, and at later instants the sequence has a probability of $0.9$ to keep the current transformations while taking a chance of $0.1$ to switch the transformation primitives. Due to the rejection sampling that excludes all-zero samples, there exists at least one active transformation primitive at every timestep; however, for datasets which are known to contain sequences without transformations, this rejection-sampling step is easily removed without any issues. For the slab component $\Tilde{\vg}_t$, the scale parameter $\lambda$ is set to $0.3$. We use the same set of hyperparameters for both datasets. 

For LatentFlow, PoFlow, and TVAE which are supervised baselines, we use their respective latent operators to move latent samples. For the unsupervised approaches, we carefully select the latent dimension that corresponds to the lowest equivariance error of a given transformation and perform a grid search to tune the traversal range in the interval $[-5,5]$. 
Since the vanilla VAE does not have any notion of learned latent transformations, when computing the equivariance error, we simply take it as a lower-bound baseline by setting the latent samples unchanged (\emph{i.e., $\vz_0=\vz_1=\dots=\vz_T$}).


\subsection{Training Details}

On MNIST~\cite{lecun1998mnist}, the training process lasts $50,000$ iterations. We only train the spike component in the first $20,000$ iterations and then integrate the slab component into the training for the rest of the iterations. On Shapes3D~\cite{3dshapes18}, we set the total training iterations to $100,000$ and split the iterations fifty-fifty into the two training stages. The batch size is set to $128$ and $64$ on MNIST and Shapes3D, respectively. We use the Adam optimizer with a learning rate of $1e{-}4$. The input images are of the size $28{\times}28$ on MNIST and of the size $64{\times}64$ on Shapes3D. The quantitative results are reported as mean $\pm$ standard deviation computer over $5$ runs with random initialization.

\begin{table*}[tbp]
    \centering
    \caption{Equivariance error $\mathcal{E}_{k}$ and log-likelihood $\log p(\vx_t)$ on MNIST using different vector fields.}
    \resizebox{0.65\linewidth}{!}{
    \begin{tabular}{c|c|c|c|c}
    \toprule
        \multirow{2}*{\textbf{Methods}}  & \multicolumn{3}{c|}{\textbf{Equivariance Error ($\downarrow$)}} & \multirow{2}*{\textbf{Log-likelihood ($\uparrow$)}} \\
        \cline{2-4}
        & \textbf{Scaling} & \textbf{Rotation} & \textbf{Coloring} & \\
        \midrule
        $\nabla u(\vz,t)$& 303.81$\pm$5.13 & 271.74$\pm$4.56 & \textbf{286.52$\pm$4.96} & -2118.49$\pm$3.09\\
         $\nabla u(\vz,t) + \vr(\vz)$& \textbf{281.32$\pm$4.71}  & \textbf{230.93$\pm$5.02} & 292.85$\pm$4.58 & \textbf{-2107.65$\pm$2.27}\\
    \bottomrule
    \end{tabular}
    }
    \label{tab:vector_field}
\end{table*}

\section{More Experimental Results}

\subsection{Disentanglement Metrics}
\label{sec:app_dis}

As discussed in~\cite{Tzelepis_2021_ICCV,song2023flow}, following $\beta$-VAE~\cite{higgins2016beta}, there are many disentanglement metrics proposed for single-dimension traversal methods, such as DCI~\cite{eastwood2018framework} and MIG~\cite{chen2018isolating}. These metrics assume that each latent dimension corresponds to one generative factor and manipulating these single dimensions would trigger distinct output transformations. The recent disentanglement methods~\cite{shen2021closed,Tzelepis_2021_ICCV,song2023householder} propose a more realistic disengagement setting: all the latent dimensions are perturbed by vector arithmetic for output variations. If one evaluates these vector-based disentanglement methods using the metrics designed for single-dimension manipulations, their scores would drop considerably and are not comparable. Nonetheless, certain disentanglement metrics such as the VP score~\cite{zhu2020learning} can be adopted for the evaluation of general disentanglement baselines as they do not pose any assumptions on the latent space. Instead, the VP metric takes a lightweight neural network to learn classifying a dataset of different image pairs $[\vx_0,\vx_T]$ under the few-shot learning setting (\emph{i.e.,} only $1\%$ or $10\%$ of the dataset is used as the training set). The validation accuracy reflects the distinguishability of these learned traversal directions, which is a reasonable surrogate for the disentanglement score. 

\begin{table}[h]
\centering
\caption{VP Scores ($\%$) on MNIST.}\label{tab:vp_mnist}
\resizebox{0.99\linewidth}{!}{
\begin{tabular}{c|c|c|c|c|c}
    \toprule
     \textbf{Training Set}  & \textbf{STA}  & \textbf{LatentFlow} & \textbf{PoFlow} & \textbf{TVAE} & \textbf{FactorVAE}   \\
     \midrule
     $10\%$ &\textbf{98.85} & {95.69} &93.05 &89.91 &85.92  \\ 
     $1\%$ & \textbf{97.04} &{92.71} &91.27 &88.15 &84.46 \\ 
    \bottomrule
\end{tabular}
}
\end{table}

\begin{table}[h]
\centering
\caption{VP Scores ($\%$) on Shapes3D.}\label{tab:vp_shapes}
\resizebox{0.99\linewidth}{!}{
\begin{tabular}{c|c|c|c|c|c}
    \toprule
     \textbf{Training Set}    & \textbf{STA}  & \textbf{LatentFlow} & \textbf{PoFlow} & \textbf{TVAE} & \textbf{FactorVAE} \\
     \midrule
     $10\%$&\textbf{97.98} &{95.92} &91.48 &88.27 &84.49 \\ 
     $1\%$&\textbf{86.09} &{77.03} &72.32 &68.39 &63.83 \\ 
    \bottomrule
    \end{tabular}
}
\end{table}

Table~\ref{tab:vp_mnist} and~\ref{tab:vp_shapes} present the quantitative evaluation of the VP scores with different split ratios of the training set on MNIST and Shapes3D, respectively. Our STA surprisingly outperforms all baselines, including both the supervised and unsupervised ones. Different from 
the supervised methods where each vector field is forced to learn one transformation, our model naturally disentangles the transformations into these learned flows through sparsity. We suspect that this gap might make our flows easier to be distinguished by small neural networks. 


\subsection{Impact of Divergence-free Vector Fields}




We leverage the Helmholtz decomposition to obtain more expressive vector fields. It is important to understand whether the extra divergence-free component could bring any concrete benefits. Table~\ref{tab:vector_field} presents the equivariance error and likelihood on MNIST using different types of vector fields. We see that adding the flow field $\vr(\vz)$ could improve the equivariance error of most transformations as well as the log-likelihood. In particular, the error of the rotation transformation is greatly improved. This again meets our expectations that the rotational field $\vr(\vz)$ improves the modeling of periodic transformations. 

\begin{table}[htbp]
    \centering
    \caption{Approximation error of each divergence-free vector field on MNIST~\cite{lecun1998mnist}.}
    \begin{tabular}{c|c|c}
     \toprule
          $||\nabla\cdot \vr^0(\vz)||_2^2$ & $||\nabla\cdot \vr^1(\vz)||_2^2$ & $||\nabla\cdot \vr^2(\vz)||_2^2$  \\
    \midrule
        0.017 & 0.032 & 0.028\\
    \bottomrule
    \end{tabular}
    \label{tab:error_div}
\end{table}

Since our divergence-free vector fields $\nabla\cdot\vr(\cdot)$ are enforced via a PINN loss, evaluating their approximation error is indeed important to verify their divergence-free property. Table~\ref{tab:error_div} reports the approximation errors for each divergence-free vector field trained on the MNIST dataset. The results indicate that our PINN effectively minimizes the divergence, achieving consistently small errors, thus validating that the learned vector fields closely approximate true divergence-free (rotational) fields.

\begin{figure*}[tbp]
    \centering
    \includegraphics[width=0.99\linewidth]{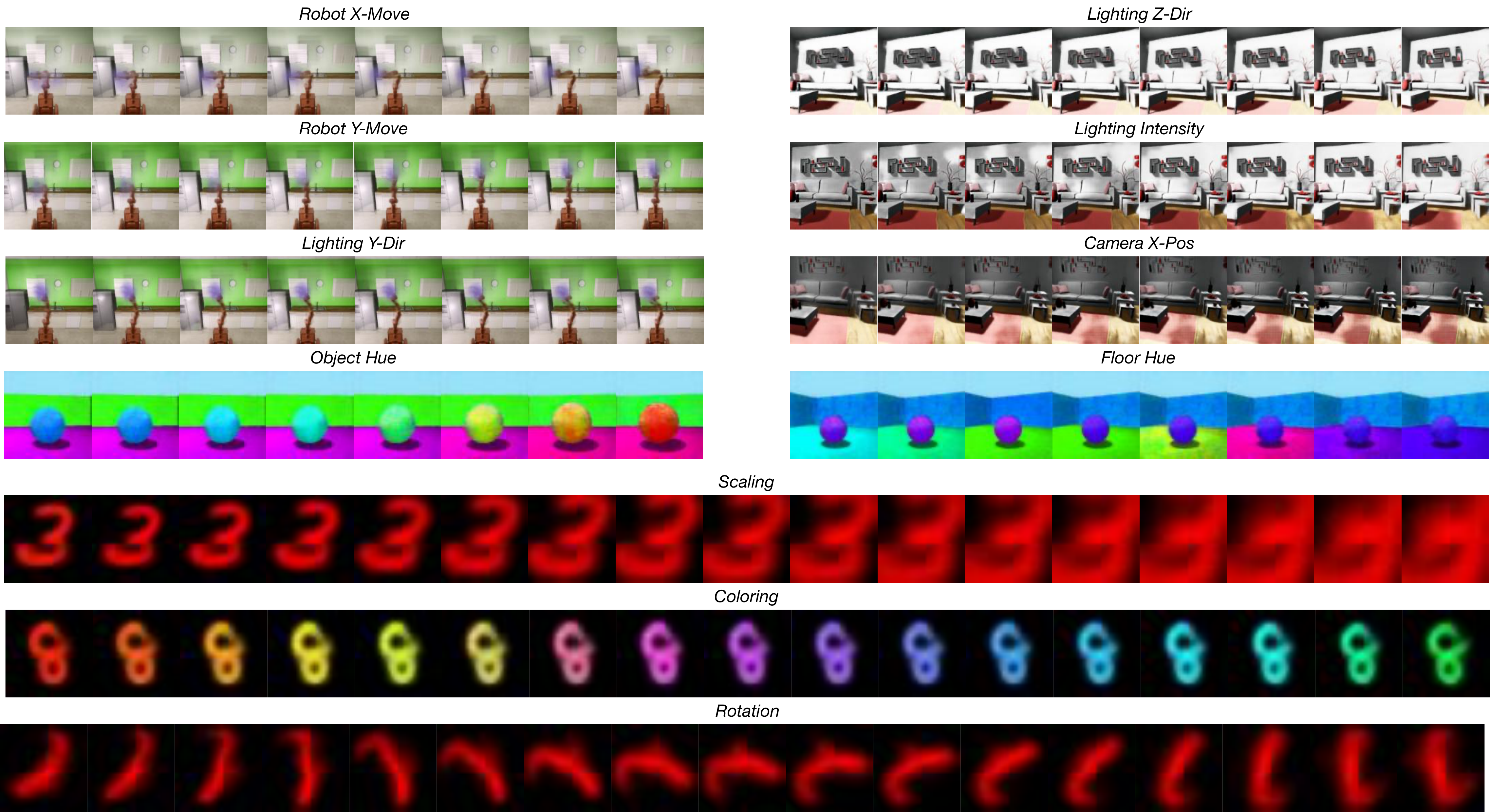}
    \caption{Traversals of our learned flow fields for more steps (longer sequences).}
    \label{fig:long_seq}
\end{figure*}

\subsection{Traversal for More Steps}

Fig.~\ref{fig:long_seq} displays the traversals of our learned flow fields for more time points (longer sequences). We see that our STA still allows for smooth interpolation across timesteps. Notice that the traversal step is even larger than the maximal sequence length of Falcol3D and Issac3D~\cite{nie2020semi}, but the latent flows keep smooth transitions during time evolution. 



\end{document}